\documentclass[11pt, letterpaper]{cmu}

\usepackage[all]{hypcap}
\usepackage[comma,numbers,sort,compress]{natbib}
\usepackage{hyperref}[citecolor=magenta]
\usepackage{amsmath, amssymb}

\hypersetup{
    colorlinks = true,
    citecolor = {magenta},
}

    \PassOptionsToPackage{numbers, compress}{natbib}
\usepackage[utf8]{inputenc} % allow utf-8 input
\usepackage[T1]{fontenc}    % use 8-bit T1 fonts
\usepackage{url}            % simple URL typesetting
\usepackage{booktabs}       % professional-quality tables
\usepackage{amsfonts}       % blackboard math symbols
\usepackage{nicefrac}       % compact symbols for 1/2, etc.
\usepackage{microtype}     % 
\usepackage{xcolor}         % colors
\microtypecontext{spacing=nonfrench}
\usepackage{amsmath,amsthm}
\usepackage{graphicx}

\usepackage{mathtools}
\usepackage{mathrsfs}
\usepackage{dsfont}
\usepackage{float}

\usepackage{amssymb}% http://ctan.org/pkg/amssymb
\usepackage{pifont}% http://ctan.org/pkg/pifont

\usepackage{xspace}
\usepackage[capitalize,noabbrev]{cleveref}
\usepackage{lipsum}
\usepackage{listings}

\usepackage{bbm}
\usepackage{tabularx}
\usepackage{lmodern}
\usepackage[T1]{fontenc}
\usepackage{lmodern}
\usepackage{amsmath, amssymb}
\usepackage{multirow}
\usepackage[export]{adjustbox}
\usepackage{colortbl}

\usepackage{graphicx}
\usepackage{makecell}   % for centered multi-line cells
\usepackage{booktabs} % for \toprule etc.
\usepackage{wrapfig}

\newcolumntype{Y}{>{\raggedright\arraybackslash}X}

\usepackage[noend]{algpseudocode} % (optional) removes "end if/while" to save space
\newcommand{\algcommentcolor}{blue!70!black}
\algrenewcommand\algorithmiccomment[1]{\hfill{\color{\algcommentcolor}$\triangleright$~#1}}

\usepackage{setspace}

\usepackage{color}
\definecolor{deepblue}{rgb}{0,0,0.5}
\definecolor{deepred}{rgb}{0.6,0,0}
\definecolor{deepgreen}{rgb}{0,0.5,0}
\definecolor{boost_correct_to_correct}{HTML}{66C2A5}
\definecolor{default_correct_to_correct}{HTML}{fc8d62}
\definecolor{dup_correct_to_correct}{HTML}{8da0cb}
\definecolor{new_correct_to_correct}{HTML}{e78ac3}

\usepackage[most,skins,theorems]{tcolorbox}

\theoremstyle{plain}

\theoremstyle{definition}

\theoremstyle{remark}

\usepackage{wrapfig}
\usepackage{subcaption}

\definecolor{rliableolive}{HTML}{BBCC33}
\definecolor{rliableblue}{HTML}{77AADD}
\definecolor{rliablered}{HTML}{EE8866}

\tcbset{
  aibox/.style={
    width=\linewidth,
    top=6pt,
    bottom=0pt,
    left=1.5pt,
    right=1.5pt,
    colback=rliableolive!13!white,
    colframe=black,
    colbacktitle=black,
    enhanced,
    center,
    attach boxed title to top left={yshift=-0.1in,xshift=0.15in},
    boxed title style={boxrule=0pt,colframe=white,},
  }
}
\newtcolorbox{AIbox}[2][]{aibox,title=#2,#1}
\usepackage{enumitem}

\newcommand\pythonstyle{\lstset{
basicstyle=\ttfamily\footnotesize,
language=Python,
morekeywords={self, clip, exp, mse_loss, uniform_sample, concatenate, logsumexp},              
keywordstyle=\color{deepblue},
emph={MyClass,__init__},          
emphstyle=\color{deepred},   
stringstyle=\color{deepgreen},
frame=single,                       
showstringspaces=false
}}

\lstnewenvironment{python}[1][]
{
\pythonstyle
\lstset{#1}
}
{}

\newcommand\pythoninline[1]{{\pythonstyle\lstinline!#1!}}

\newcommand{\methodname}{\emph{RaC}}

\definecolor{blanchedalmond}{rgb}{1.0, 0.92, 0.8}
\definecolor{carmine}{rgb}{0.59, 0.0, 0.09}
\definecolor{lightblue}{rgb}{0.22,0.45,0.70}% light blue

\newcommand{\termbold}[1]{{\color{lightblue} \emph{\textbf{#1}}}}

\renewcommand{\mathbf}{\boldsymbol}

\makeatletter
\def\Ddots{\mathinner{\mkern1mu\raise\p@
\vbox{\kern7\p@\hbox{.}}\mkern2mu
\raise4\p@\hbox{.}\mkern2mu\raise7\p@\hbox{.}\mkern1mu}}
\makeatother
\numberwithin{equation}{section}

\definecolor{amaranth}{rgb}{0.9, 0.17, 0.31}
\definecolor{antiquebrass}{rgb}{0.8, 0.58, 0.46}
\definecolor{antiquefuchsia}{rgb}{0.57, 0.36, 0.51}
\definecolor{chromeyellow}{rgb}{0.31, 0.47, 0.26}

\usepackage[dvipsnames]{xcolor}
\definecolor{maj5}{HTML}{2b8cbe}
\definecolor{maj5Imp}{HTML}{084081}

\definecolor{seq5wo}{HTML}{d95f0e}
\definecolor{seq5woImp}{HTML}{662506}

\definecolor{seq5w}{HTML}{6a51a3}
\definecolor{seq5wImp}{HTML}{3f007d}

\definecolor{selfwo}{HTML}{d95f0e}
\definecolor{selfwoImp}{HTML}{662506}

\definecolor{selfw}{HTML}{6a51a3}
\definecolor{selfwImp}{HTML}{3f007d}

\definecolor{glorewo}{HTML}{d95f0e}
\definecolor{glorewoImp}{HTML}{662506}

\definecolor{glorew}{HTML}{6a51a3}
\definecolor{glorewImp}{HTML}{3f007d}

\definecolor{vstar}{HTML}{d95f0e}
\definecolor{vstarImp}{HTML}{662506}

\makeatletter
\def\mathcolor#1#{\@mathcolor{#1}}
\def\@mathcolor#1#2#3{%
  \protect\leavevmode
  \begingroup
    \color#1{#2}#3%
  \endgroup
}
\makeatother

\usepackage[textsize=tiny]{todonotes}
\usepackage{algorithm}
\usepackage{amssymb}

\usepackage[skins,theorems]{tcolorbox}
\Crefformat{equation}{#2Eq.\;(#1)#3}

\Crefformat{figure}{#2Figure #1#3}
\Crefformat{assumption}{#2Assumption #1#3}
\Crefname{assumption}{Assumption}{Assumptions}

\usepackage{crossreftools}
\usepackage[suppress]{color-edits}

\usepackage{dsfont}
\usepackage{nicefrac}

\newtcolorbox{analysisbox}[1][]{
    enhanced jigsaw,
    colback=white,
    colframe=blue!75!black,
    fonttitle=\bfseries,
    boxsep=5pt,
    left=5pt,
    right=5pt,
    top=5pt,
    bottom=5pt,
    title=#1,
}

\definecolor{rliableolive}{HTML}{BBCC33}
\definecolor{rliableblue}{HTML}{77AADD}
\definecolor{rliablered}{HTML}{EE8866}

\tcbset{
  aibox/.style={
    width=\linewidth,
    top=8pt,
    bottom=4pt,
    colback=rliableolive!8!white,
    colframe=black,
    colbacktitle=black,
    enhanced,
    center,
    attach boxed title to top left={yshift=-0.1in,xshift=0.15in},
    boxed title style={boxrule=0pt,colframe=white,},
  }
}
\definecolor{lightblue}{rgb}{0.22,0.45,0.70}% light blue

\usepackage{pifont}
\usepackage{soul}
\definecolor{highlightmistake}{RGB}{255, 179, 179}
\definecolor{highlightcorrect}{RGB}{179, 255, 179} 

\title{\methodname{}: Robot Learning for Long-Horizon Tasks by Scaling \underline{R}ecovery \underline{a}nd \underline{C}orrection}

\author[1]{Zheyuan Hu}
\author[1]{Robyn Wu}
\author[1]{Naveen Enock}
\author[1]{Jasmine Li}
\author[1]{Riya Kadakia}
\author[1]{Zackory Erickson$^\star$}
\author[1]{Aviral Kumar$^\star$}

\affil[1]{Carnegie Mellon University}

\correspondingauthor{zheyuanh@andrew.cmu.edu. ~~ $^\star$Co-advising.}

\begin{abstract}
\textbf{Abstract:} 
Modern paradigms for robot imitation train expressive policy architectures on large amounts of human demonstration data. Yet performance on contact-rich, deformable-object, and long-horizon tasks plateau far below perfect execution, even with thousands of expert demonstrations. This is due to the inefficiency of existing ``expert'' data collection procedures based on human teleoperation. To address this issue, we introduce \methodname{}, a new phase of training on human-in-the-loop rollouts after imitation learning pre-training. In \methodname{}, we fine-tune a robotic policy on human intervention trajectories that illustrate recovery and correction behaviors. Specifically, during a policy rollout, human operators intervene when failure appears imminent, first rewinding the robot back to a familiar, in-distribution state and then providing a corrective segment that completes the current sub-task. Training on this data composition expands the robotic skill repertoire to include retry and adaptation behaviors, which we show are crucial for boosting both efficiency and robustness on long-horizon tasks. Across three real-world bimanual control tasks: shirt hanging, airtight container lid sealing, takeout box packing, and a simulated assembly task, \methodname{} outperforms the prior state-of-the-art using 10$\times$ less data collection time and samples. We also show that \methodname{} enables test-time scaling: the performance of the trained \methodname{} policy scales linearly in the number of recovery maneuvers it exhibits. Videos of the learned policy are available at \href{https://rac-scaling-robot.github.io/}{https://rac-scaling-robot.github.io/}.
\end{abstract}

\begin{document}

\maketitle

\vspace{-0.3cm}
\section{Introduction}
\label{sec:introduction}
\vspace{-0.2cm}

Running imitation learning with expressive models on human teleoperation data powers a large chunk of modern robotic learning. In fact, a number of recent academic and industrial bets have been on massively scaling up imitation learning as a form of pre-training for robots~\citep{black2024pi0visionlanguageactionflowmodel, trilbmteam2025carefulexaminationlargebehavior, bu2025agibot, octo_2023, rt12022arxiv, rt22023arxiv, geminiroboticsteam2025geminiroboticsbringingai, zhao2024alohaunleashedsimplerecipe}.
However, results increasingly suggest that this paradigm is approaching a performance ceiling well below perfect task completion. For example, even with over $5000$ human demonstrations, state-of-the-art task-specific models can only place a single t-shirt on a hanger with bimanual manipulators at roughly $75\%$ success. While one might hope that more data or alternative learning frameworks could close this gap, in practice these methods still struggle to overcome compounding errors and stochasticity in long-horizon tasks.

We argue that this limitation of imitation is fundamental: while mimicking expert actions can imbue the policy with ``basic'' useful skills, doing so is inherently suboptimal when the robot faces task variations or new initial states, the environment is stochastic or noisy, or the task is inherently long-horizon, where failing at one stage inhibits success in the rest (i.e., when ``compounding errors'' can be catastrophic)~\citep{ghosh2021generalization}. 
% (i.e., when ``compounding errors'' can be problematic).
As a result, policies trained via imitation learning often fail to generalize to real-world stochasticity and dynamism, and exhibit diminishing returns with additional data, leading to a performance plateau. Crucially, this failure stems not from the learning algorithm or the model but from the data distribution itself: demonstrations are biased toward clean, successful trajectories, but do not imbue the policy with behaviors needed to tackle compounding errors stemming from stochasticity in long-horizon tasks. 

In this work, we propose an alternative paradigm for training robot policies that directly addresses the limitations of success-only imitation learning. We introduce a new phase of learning that is run subsequent to basic imitation learning  on clean teleoperation data (``pre-training''), which we refer to as \textbf{{\methodname{}}}. The central idea of \methodname{} is to train on trajectories that interleave successful task executions with segments that demonstrate recovery, retries, and adaptation: behaviors that are essential for robustness in complex or novel situations. While standard human teleoperation data may already contain some incidental recovery behavior,\footnote{For instance, in a study of the DROID dataset, we find that only $3.68\%$ of the episodes contain recovery behavior.} \methodname{} explicitly encourages and amplifies such behaviors. Conceptually, this phase is analogous to ``mid-training'' for large language model (LLM) reasoning~\citep{wang2025octothinkermidtrainingincentivizesreinforcement}, which aims to illustrate how to best combine basic knowledge with algorithmic behavior (e.g. backtracking, trial-and-error, self-verification, etc.) to solve complex reasoning problems by producing much longer responses.

\begin{figure}[t]
  \centering
  \vspace{-0.3cm}
  \includegraphics[width=0.83\linewidth]{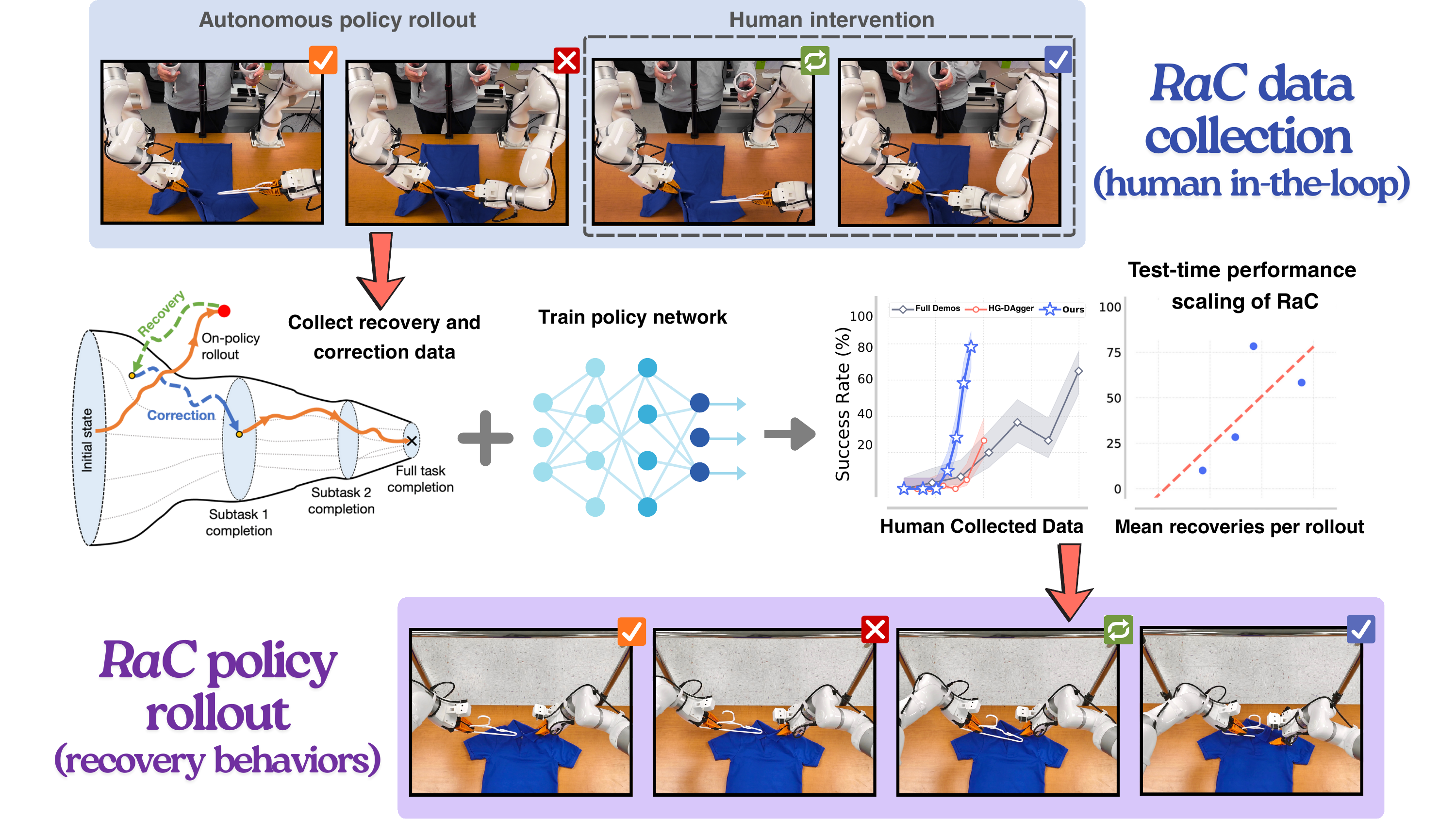}
  \vspace{-0.25cm}
  \caption{\footnotesize{\emph{\textbf{Illustrating \methodname{}.}} Our approach enables imitation learning policies to robustly execute long-horizon tasks by explicitly learning skills such as recovery and correction to handle mistakes and failures. Doing so substantially improves data efficiency and results in effective performance scaling at test time by executing rollouts with more recovery maneuvers.}}
  \label{fig:teaser}
  \vspace{-0.4cm}
\end{figure}

Concretely, we introduce a lightweight \textbf{human-in-the-loop} data collection protocol: human teleoperators intervene to take control from the running policy as soon as it begins to deviate from the expected course. As shown in Figure \ref{fig:recovery_figure}, these interventions naturally fall into two categories: \emph{a) error correction segments}, where human experts guide the robot to solving tasks (similar in spirit to DAgger-style supervision), and \emph{b) recovery segments}, where the human rewinds or repositions the robot to a previously successful state.
To scale up recovery and correction for imitation learning, \methodname{} standardizes interventions with two rules. \emph{Rule 1 (recover then correct)} structures every human takeover into a reset back to in-distribution states followed by a corrective segment that completes the current sub-task. \emph{Rule 2 (termination after intervention)} ends the episode immediately once the intervention segment finishes, which avoids collecting data on later sub-tasks under state distributions from a mixture of learned policy and human expert. Crucially, \methodname{} keeps the imitation objective unchanged; performance gains come purely from improved data composition. Applied to three challenging, long-horizon real-world bimanual control tasks, shirt hanging, airtight-lid sealing, and clamshell takeout-box packing, \methodname{} outperforms batched full-demonstration and HG-DAgger style human-in-the-loop collection, both in performance and in data efficiency. In particular, \methodname{} achieves higher success rates and steeper scaling trends than batched full demonstration and HG-DAgger-style human-in-the-loop data collection, demonstrating superior data efficiency up to 1 order of magnitude. We further show that, analogous to long chain-of-thought (CoT) reasoning in language models~\citep{guo2025deepseek,openai_learning_to_reason_2024}, policies trained with \methodname{} exhibit \textbf{\emph{test-time scaling}} (Figure~\ref{fig:lid_retries_vs_success}): as the deployed policy executes more recovery maneuvers at test-time, its overall task success rate improves as it can try multiple times.

\textbf{Contributions.} We introduce \methodname{}, a framework for scaling imitation learning in long-horizon manipulation by leveraging \emph{recovery and correction}. Real-robot experiments show that conventional data pipelines lack the ingredients to learn diverse skills needed to handle out-of-distribution states, which arise frequently due to compounding errors. Our approach delivers test-time scaling benefits akin to ``o1-style'' LLMs, absent in prior work. On three challenging real-world bimanual tasks, \methodname{} achieves higher success rates and steeper scaling trends than full demonstrations or HG-DAgger-style interventions.

\vspace{-0.35cm}
\section{Related Work}
\label{sec:related}
\vspace{-0.2cm}

\emph{\textbf{Scaling data in robotic learning.}} Recent work shows that scaling real-robot data across tasks, embodiments, and environments enables generalization. Large robotic datasets~\citep{khazatsky2024droid, walke2023bridgedata, open_x_embodiment_rt_x_2023, bu2025agibot},
paired with highly expressive neural network architectures~\citep{rt12022arxiv, rt22023arxiv, octo_2023, kim24openvla, black2024pi0visionlanguageactionflowmodel, liu2024rdt, trilbmteam2025carefulexaminationlargebehavior},
have produced \emph{generalist} policies that achieve strong performance on many atomic skills (e.g., grasping an object, folding cloth). In parallel, another line of work~\citep{zhao2024alohaunleashedsimplerecipe, chi2024universal} demonstrates that a similar data-driven recipe can also produce \emph{specialist} policies that perform very well on substantially more complex dexterous bimanual tasks. However, these approaches require collecting \emph{thousands} of high-quality expert demonstrations per skill~\citep{zhao2024alohaunleashedsimplerecipe}.  

\emph{\textbf{Scaling studies in robot imitation learning.}} Inspired by work in LLMs~\citep{kaplan2020scaling,hoffmann2022training}, several works aim to build scaling laws for robotic imitation~\citep{zha2025guidingdatacollectionfactored, gao2024efficientdatacollectionrobotic, lin2024data}. Aimed at evaluating generalization across variations in the task, some of these works analyze the performance of policies on short-horizon tasks as a function of the environmental diversity present in the training data.
However, in all such studies, the demonstrations themselves are collected via human ``expert'' teleoperation and exhibit little variation within the sorts of skills shown in the data. In contrast, instead of environment diversity, we focus on the data collection strategy \emph{within} a {trajectory} for long-horizon tasks: specifically, the kinds of maneuvers, recovery behaviors, and variations within. As we show in our experiments, carefully designing a trajectory-level collection strategy can improve efficiency by more than $10\times$ compared to previous work~\citep{zhao2024alohaunleashedsimplerecipe} with similar tasks.

\emph{\textbf{Human-in-the-loop imitation learning.}}
Our approach collects intervention data by emphasizing recovery and correction behaviors, which connects it to the broad literature on human-in-the-loop imitation learning. Classical approaches are rooted in DAgger~\citep{pmlr-v15-ross11a}, which alternates between \textbf{1)} running on-policy rollouts from the learner, \textbf{2)} querying the expert on visited states, and \textbf{3)} retraining on the aggregated dataset. This framework assumes access to a high-quality expert policy. To adapt DAgger to human operators, HG-DAgger~\citep{kelly2019hg} enables teleoperators to provide interventions when policy visits undesirable states, while more recent systems such as RoboCopilot~\citep{wu2025robocopilothumanintheloopinteractiveimitation} extend these ideas to bimanual mobile manipulation by developing improved interfaces for teleoperation and intervention. Other works~\citep{mandlekar2020human, liu2022robot} explore objectives that combine on-policy rollouts, intervention data, and full human demonstrations. Although our learning objective bears similarities to HG-DAgger~\citep{kelly2019hg}, we depart from its formulation in a crucial way: prior works largely treat human intervention as an optimal expert solution to be imitated but we show that collecting recovery segments, which by themselves are not task-optimal (and may even undo progress on a subtask), yields substantially better scaling. This challenges the conventional wisdom that only ``expert'' interventions are useful and highlights the role of trajectory-level data collection.

\emph{\textbf{Shared autonomy.}} Effectively collecting intervention data requires responsive and intuitive teleoperation interfaces. Prior human-in-the-loop systems have typically relied on 6-DoF SpaceMouse~\citep{liu2022robot, 10610040, luo2025precisedexterousroboticmanipulation} or smartphone softwares with on-screen buttons and IMU sensing~\citep{mandlekar2020human}. While functional, these devices come with steep learning curves\citep{li2025train} and are difficult to use for dexterous skills, particularly those requiring wrist rotation. As a result, they are mostly limited to single-arm settings or relatively simple manipulation tasks where end-effector poses remain constrained. More recent work~\citep{wu2025robocopilothumanintheloopinteractiveimitation} has explored combining VR joysticks with exoskeleton hardware to provide force feedback and richer intervention options, but this demands specialized equipment and additional cost. In contrast, we adopt widely available off-the-shelf VR joysticks as our teleoperation and intervention interface. With a lightweight software modification that we described in Section~\ref{subsec:shared_autonomy_interface}, our design enables users to take over control and provide interventions instantly, without the need to align the VR joystick poses with the robot end effector poses.

\textbf{\emph{Recovery and correction in imitation learning.}} Several works also study employing recoveries and corrections for training via imitation learning. \citet{wang2025geniecenturionacceleratingscalable} proposes a ``rewind-and-refine'' data collection system that detects failures, returns the robot to a previous pose via replaying the trajectory, and then the teleoperator collects corrective trajectories. Similarly, \citep{ankile2024juicerdataefficientimitationlearning, hoque2024intervengeninterventionaldatageneration} studies generating and filtering recovery trajectories automatically in simulation to augment dataset coverage. However, these works are restricted to pure simulation tasks or limited sim2real settings. \citet{sun2025latentpolicybarrierlearning} trains a base diffusion policy on expert data and a learned latent dynamics model that performs test-time steering, encouraging the policy to stay on the expert demonstration manifold. \citet{ke2023ccil} learns a locally Lipschitz dynamics model from expert demonstrations and synthesizes corrective labels near the demo manifold to mitigate compounding errors. \citet{xu2025compliantresidualdaggerimproving} combines a compliant intervention interface to provide corrections and learns a residual policy to improve the performance of the contact-rich tasks. Instead of engineering the return to in-distribution states through an engineered rewind mechanism or modifications to the base imitation learning policy, our approach \methodname{} treats recovery as yet another ``skill'' to learn from human demonstrations and scales it explicitly alongside full demonstration and correction skills. Hence, without modifying existing imitation learning objectives or adding additional complexity to the robot system, we improve the robustness and performance of the policy by directly scaling human demonstration data. \citet{10161096} proposes a somewhat similar data collection protocol to \methodname{}, in which operators deliberately collect sequences of visually similar failures, recoveries, and successes by backtracking to earlier visual states. However, \citet{10161096} studies the benefit of such data collection strategy through the lens of offline reinforcement learning, enabling efficient learning of accurate value functions from small datasets. We instead focus on scaling properties of such recovery skills in dataset composition and their impact on imitation learning policy.

\vspace{-0.3cm}
\section{Background and Robot Setup}
\label{sec:prelim}
\vspace{-0.15cm}

\begin{wrapfigure}{r}{0.4\textwidth}
  \centering
  \vspace{-1.0cm}
  \includegraphics[width=\linewidth]{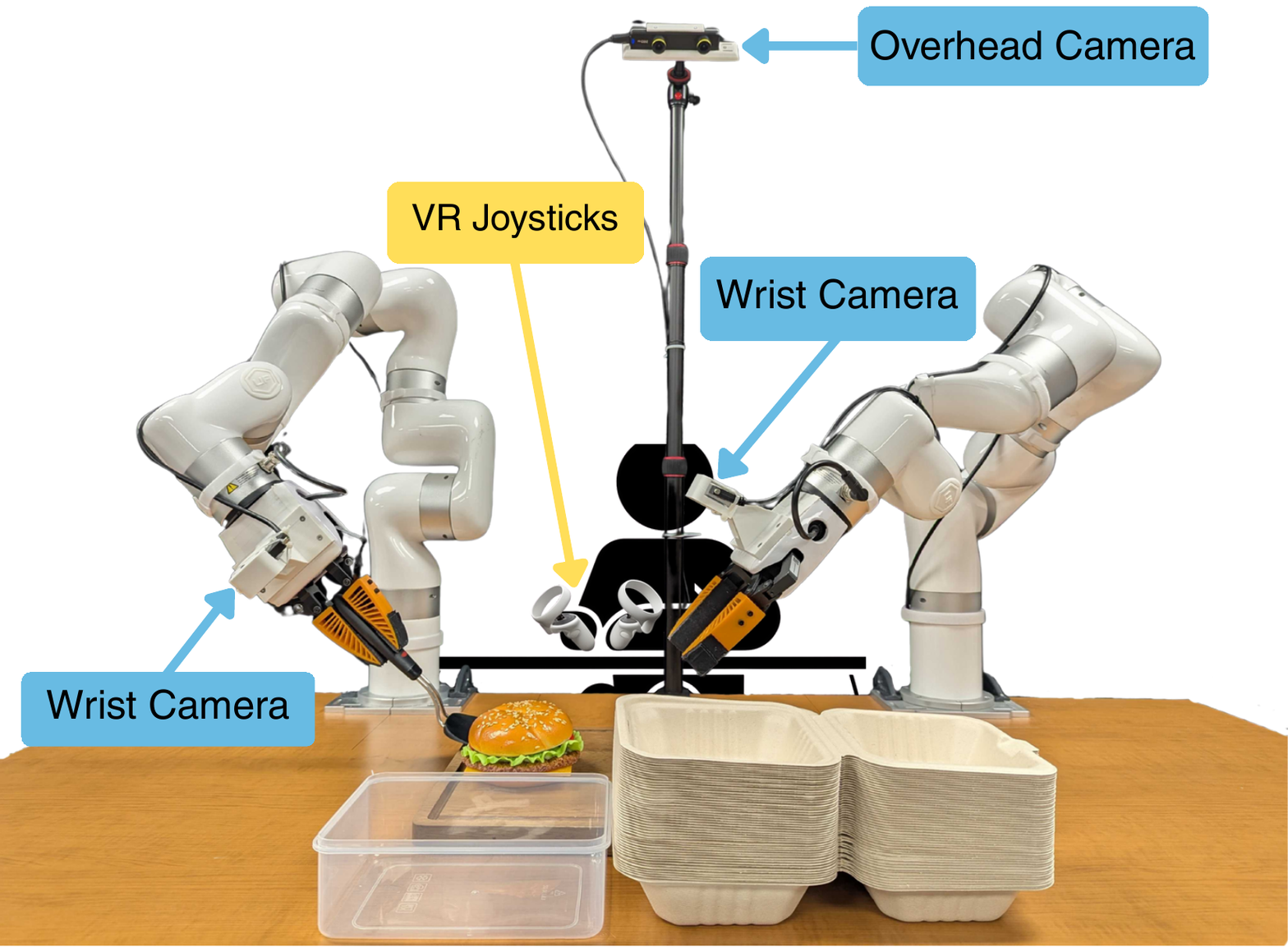}
  \vspace{-0.75cm}
  \caption{\footnotesize{\textbf{\emph{Bimanual manipulation robot system.}} An illustration of our bimanual robot setup showing camera placements and workspace setup.}}
  \label{fig:robot_setup}
  \vspace{-0.8cm}
\end{wrapfigure}

\textbf{Robot setup.} Our robot setup (Figure~\ref{fig:robot_setup}) consists of two 7-DoF xArm-7 manipulators with scaled-down version of soft grippers \citep{chi2024universal, zhaxizhuoma2025fastumiscalablehardwareindependentuniversal} to facilitate contact-rich and dexterous tasks. To obtain reactive control, a central server synchronizes and publishes RGB image streams from a top-view camera and two wrist cameras, robot state, and action commands at 60Hz. Our system utilizes RMPFlow~\citep{cheng2019rmpflowcomputationalgraphautomatic} as the inverse-kinematic motion generator, enabling real-time collision avoidance and smooth arm motions. 

From a purely machine learning standpoint, our work is situated in the setting of iterative imitation learning with evolving robotic datasets. Each trajectory $\tau$ in this dataset consists of an action $a_t$ for every observation $s_t$. In this paper, we develop an approach to collect data for imitation learning that results in better scaling by incorporating human interventions on a previous generation of the learned policy. 
Formally, our goal is to develop an \emph{iterative human data collection strategy} that improves scaling of task performance as a function of data-collection budget. In other words, we aim to improve the \emph{scaling behavior}, i.e., the slope of task success rate vs.\ data size. To study data compositions, our data consists of three types: \textbf{(i)} full, successful expert demonstrations $\mathcal{D}^{\mathrm{full}}$; \textbf{(ii)} \emph{recovery} segments, that begin in failure or out-of-distribution regions and return to in-distribution regions; and \textbf{(iii)} \emph{correction} segments that directly complete the current subtask.

\textbf{Data collection protocol.} Our data collection begins with collecting one round of full demonstration data using an initial budget size $R_0$, in terms of hours or the number of frames/timesteps. We then first train an initial policy $\pi_{0}$ using this ``Round 0'' full-demonstration data and evaluate its performance. Round 0 data could also come from off-the-shelf imitation learning dataset already available in the community. We can then scale data by following two protocols. In the \emph{batched data collection} protocol, we allocate an additional budget of $K \times R_0$ frames, yielding a single batch of expert data of size $(K+1) \times R_0$. In the \emph{iterative human intervention} protocol~\citep{kelly2019hg, wu2025robocopilothumanintheloopinteractiveimitation, mandlekar2020human, liu2022robot}, experts instead perform $K$ alternating rounds of intervention and training: in each round $k$, they provide interventions during rollouts of $\pi_{k-1}$, aggregate these intervention segments with existing data (in different ways), and retrain a new policy $\pi_k$. We will study the nature of interventions that improve data scaling of imitation learning the most.

\begin{figure}[t]
  \centering
  \vspace{-0.3cm}
  \includegraphics[width=0.8\linewidth]{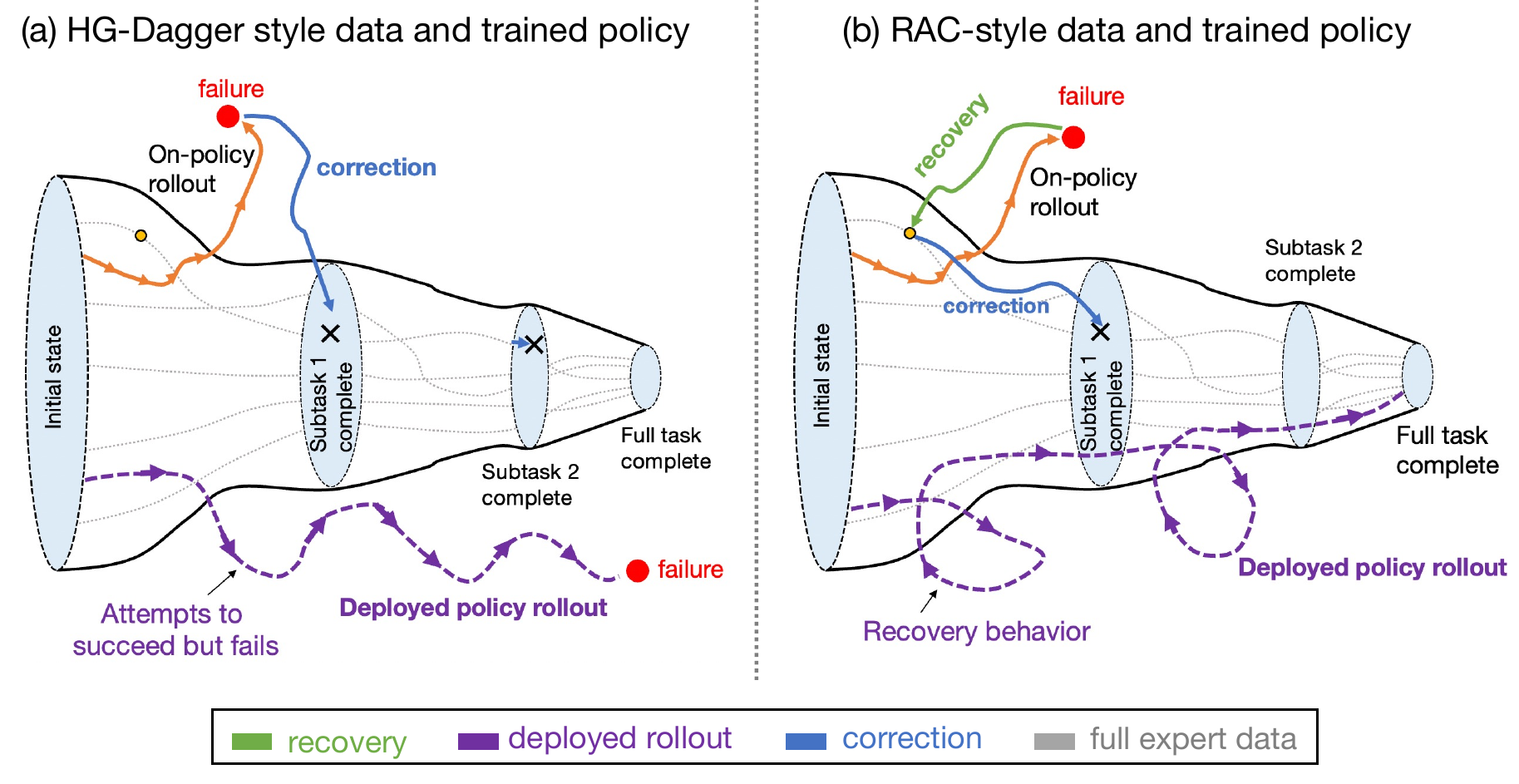}
  \vspace{-0.25cm}
  \caption{\footnotesize{\textbf{\emph{Illustrating the core concept behind \methodname{}}.} Data collected via human interventions prescribed by \methodname{} and a sample policy rollout when training on only correction data (``HG-DAgger'') vs recovery and correction data (\methodname{}). Typical intervention approaches (left) simply collect correction data that pushes the task forward from out-of-distribution states, aiming to push the task forward. In contrast, \methodname{} (right) collects a recovery segment that places the robot into a prior familiar state followed by a correction segment that pushes the task forward from this state. This ``densifies'' coverage over familiar states and teaches the robot to also recover to a broader region of initial states. Since learning to recover is easier than pushing the sub-task forward, \methodname{} exploits a form of ``verification-generation gap''~\citep{setlur2025scaling}: it allows the policy to succeed more by recovering multiple times.}}
  \label{fig:recovery_figure}
  \vspace{-0.35cm}
\end{figure}

\vspace{-0.3cm}
\section{\methodname{}: Scaling \underline{\textit{R}}ecovery \underline{\textit{a}}nd \underline{\textit{C}}orrection for Imitation Learning}
\vspace{-0.15cm}
Our goal is to design an iterative data collection strategy for scaling imitation learning. Unlike prior iterative approaches that collect corrective segments~\cite{kelly2019hg, wu2025robocopilothumanintheloopinteractiveimitation,mandlekar2020human,liu2022robot}, our approach deliberately guides human interventions to include a substantial proportion of ``recovery'' segments alongside ``corrective'' segments. While recovery segments are \emph{suboptimal} for completing any sub-task within the long-horizon task, they bring the policy back into an in-distribution state preemptively, giving it a chance to re-attempt the task (Figure~\ref{fig:recovery_figure}). In contrast, corrective segments illustrate how to complete the task. \emph{\textbf{Our main insight}} is that the ability to retry multiple times gives the policy a generic recipe to attenuate compounding errors that often bottlenecks imitation, by trading off acting longer for lower error. We formalize this notion of recovery and correction and develop an approach to collect data naturally rich in these behaviors.

\vspace{-0.25cm}
\subsection{Understanding the Role of Recovery and Correction Segments in Imitation}
\vspace{-0.2cm}

Consider a robot policy $\pi$ that executes a trajectory $\tau = (s_0, a_0, s_1, a_1, \ldots, s_t)$, where $s_t$ denotes the state at which a human expert intervenes. A sequence of human actions $(a^\text{h}_{t+1}, a^\text{h}_{t+2}, \ldots, a^\text{h}_{t+k})$ starting from $s_t$ constitutes a \emph{\textbf{recovery segment}} if the resulting state $s^\text{h}_{t+k}$ that the robot reaches after the intervention lies within the distribution of states visited in the prefix of human demonstrations $\mathcal{D}^\text{full}[0:t]$. Conversely, this sequence of actions constitutes a \textbf{corrective segment} if the resulting state $s^\text{h}_{t+k}$ lies within the distribution of states visited after timestep $t$ in demonstrations $\mathcal{D}^\text{full}[t+1:H]$. We illustrate this concept in Figure \ref{fig:recovery_figure}.

\textcolor{lightblue}{\textbf{\emph{How can recovery segments improve performance?}}} Intuitively, recovery segments return the policy to familiar previous states, giving it another chance to attempt the task, whereas corrective segments show how to push the trajectory forward. This raises a question: can a policy actually learn to ``reset'' itself by imitating recovery segments, and why would this improve performance? \textbf{Our key intuition} is that in tasks where the set of valid initial states is broad (e.g., for the task of inserting a t-shirt on an hanger, any configuration where a shirt lies on a table and a hanger is in one of the robot's arms somewhere above the shirt is an initial state) but the set of valid goal states is narrow (e.g., only when the shirt is correctly inserted on the hanger resting on the rack), resetting to a previously encountered state is generally far easier than executing a sub-task correctly (e.g., inserting the collar of the shirt onto the hanger). Because there are multiple familiar past states to reset to, recovery requires less precision and can be more sample-efficient to learn than solving the task. This is akin to the presence of a verification-generation gap (VG-gap)~\citep{setlur2025scaling}, where learning one skill (recovery) is more sample-efficient than the other.\footnote{For most long-horizon tasks in the real world, this structure naturally arises: progress on earlier sub-tasks is often essential for success on later ones. Furthermore, because large-scale imitation learning systems typically aggregate demonstrations from multiple teleoperators, the resulting data introduces substantial diversity, especially early-on in the attempt to solve the task.}

This means that training via imitation learning on a mixture of recovery and corrective behavior should equip a policy with two complementary ways to improve performance: \textbf{(1)} by mimicking corrective segments (and full demonstration) to make progress in the first shot, with at least some probability, and \textbf{(2)} by resetting to a previous familiar state and retrying. When the setting exhibits the structure above, this ability to recover can be acquired with relatively little data. Once the policy can reliably recover from an anticipated failure, repeated retries would then naturally amplify the overall probability of producing at least one attempt that correctly executes the sub-task. In fact, the probability of never succeeding on a sub-task decays exponentially with the number of retries. This means that total suboptimality in imitation learning performance should decrease.  This mechanism is akin to sequential test-time scaling~\citep{snell2024scaling} in large language models (LLMs): just as long chain-of-thought (CoT) models~\citep{guo2025deepseek} improve performance and generalization by spending more tokens on backtracking and recovery before re-attempting a question, we expect \methodname{} to achieve similar gains by performing backtracking and retrying directly in action space. 

\textcolor{lightblue}{\textbf{\emph{How can recovery segments improve \underline{data scaling} relative to HG-DAgger style methods?}}} Recovery segments improve data efficiency because returning to familiar in-distribution states requires less data than mastering corrective sub-tasks in many cases. From in-distribution states, the policy already has strong supervision from existing data and the newly added corrective segments amplify this supervision. In contrast, methods like HG-DAgger demonstrate an entirely new behavior from an unfamiliar out-of-distribution state and require the policy to master it. As a result, performance as a function of data scale is expected to be lower for HG-DAgger since it does not necessarily amplify coverage over either in-distribution states or new unfamiliar states within limited intervention budgets. 

\begin{AIbox}{Summary: Recovery and Correction Behaviors for Imitation}
\begin{itemize}[itemsep=2pt]
 \setlength{\leftskip}{-10pt}
    \item Simply scaling full demonstrations from experts prioritizes optimal trajectories but leaves \emph{out-of-distribution (OOD)} and failure states under-covered, leading to compounding error.
    \item Training on trajectories that recover from failure or OOD states and complete task after recovery promotes recovery-then-retry behaviors, boosting performance and data efficiency.
\end{itemize}
\end{AIbox}

\vspace{-0.25cm}
\subsection{Scaling Recovery and Correction Segments in Human Teleoperation}
\label{subsec:rac_rules}
\vspace{-0.15cm}

Next, we turn to the question of how to collect imitation data that contains a substantial proportion of both recovery and correction segments. In principle, one could simply \emph{instruct} human teleoperators to artificially stage possible failure states, and demonstrate recovery and corrective behaviors~\citep{10161096}. However, such behaviors produced by humans from contrived or ``fake'' states may not reflect the out-of-distribution errors that a learned policy would actually encounter. Since policy mistakes are tightly coupled with the policy itself, a purely \emph{offline} approach is unlikely to be effective (akin to LLMs~\citep{setlur2025scaling}). A more effective alternative is to collect this data through \emph{human-in-the-loop} interventions. Analyzing human intervention data in Section \ref{subsec:learn_dynamics}, we find that it is difficult to achieve a good balance between recovery and correction with no standardization of data collection protocol. Thus, we prescribe two simple rules for intervention:

\underline{\termbold{\textbf{\emph{Rule 1: Pair each recovery segment with a correction segment.}}}} Each intervention is structured to contain two phases. First, the human operator performs \emph{recovery} behavior by executing a sequence of actions that bring the robot system back into a familiar in-distribution region of states. Then, the operator provides \emph{corrective} behavior, attempting to push the current sub-task forward (see Figure \ref{fig:recovery_figure} for an illustration). This simple structure ensures that every intervention teaches the policy both how to reset itself and how to make progress, rather than overemphasizing one or the other.

\underline{\termbold{\textbf{\emph{Rule 2: Terminate after intervention.}}}} After an intervention concludes, we terminate the entire episode. In long-horizon tasks, later sub-tasks depend on the correct execution of earlier ones. Allowing the rollout to continue after human intervention would contaminate later sub-tasks with a distribution of states induced by a combination of the learned policy and the human teleoperator. While not problematic in itself, learning on this distribution of states might not necessarily improve the policy under its \emph{own} induced distribution of states when it attempts the later sub-tasks, which can be fairly different from the joint human and policy distribution in a particular intervention rollout. 
This means that intervention data for later sub-tasks after an intervention is likely to not help much, but cost us more samples. Instead, terminating early allows allocating a total data collection budget more towards improving early sub-tasks.

% \vspace{-0.1cm}
\begin{AIbox}{Summary: Balanced composition of recovery and correction}
For the widely-used DROID dataset~\citep{khazatsky2024droid}, an analysis on its 1\% sub-sample reveals only $3.68\%$ of episodes contain $\geq 1$ recovery and $16.58\%$ contain $\geq 1$ correction. Similarly, in Section \ref{subsec:learn_dynamics}, our HG-DAgger data skews heavily toward corrections with scarce recovery segments. \methodname{} standardizes data collection for interventions: pair a recovery with a correction, then terminate, to produce a balanced mixture of skills, improving robustness and data efficiency.
\end{AIbox}

\subsection{Shared Autonomy Interface and Guidance for Teleoperators}
\label{subsec:shared_autonomy_interface}
% \vspace{-0.15cm}

\vspace{-0.15cm}
\begin{figure}[h]
  \centering
  \vspace{-0.15cm}
  \includegraphics[width=0.8\linewidth]{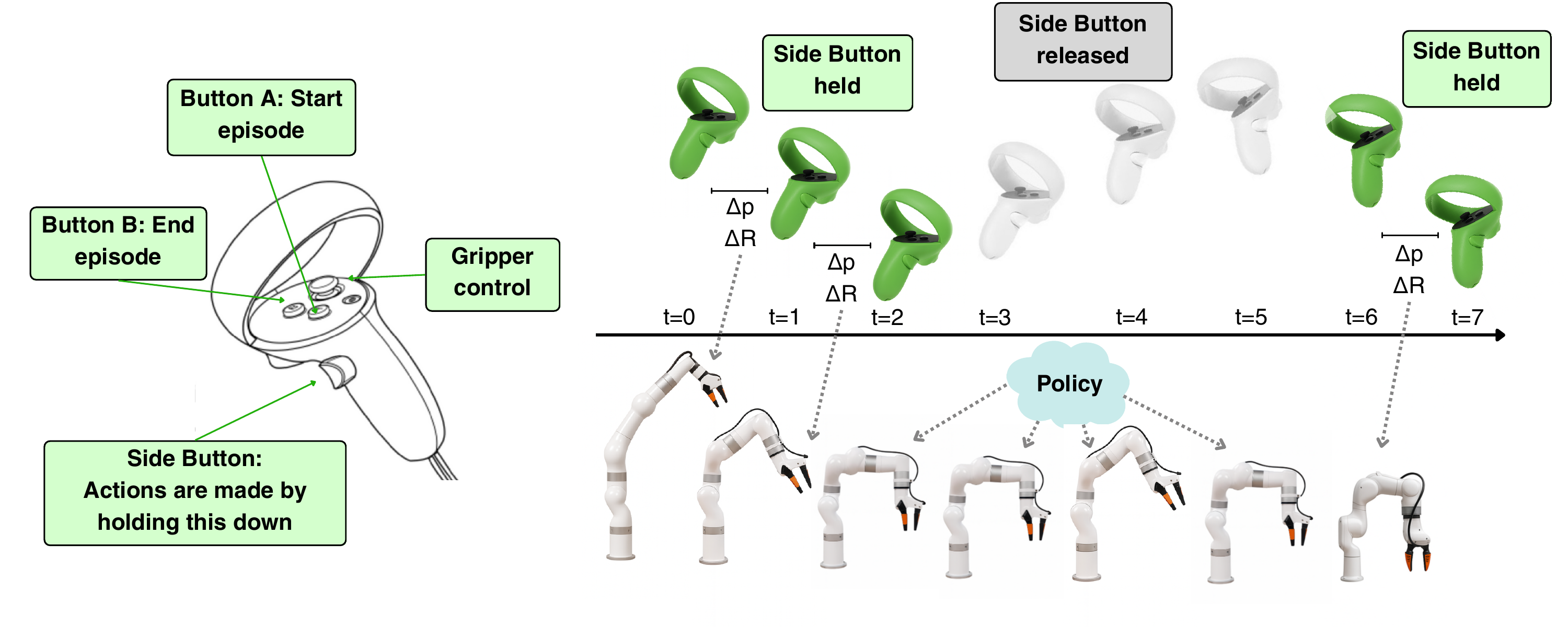}
  \vspace{-0.2cm}
  \caption{\footnotesize{\emph{\textbf{VR handset interface for shared autonomy in \methodname{}.}} We design and implement a ``clutch'' design that enables smooth handover from the robot policy to the human teleoperator.}}
  \label{fig:vr_handset}
  \vspace{-0.2cm}
\end{figure}
To enable effective interventions for \methodname{}, we design a lightweight shared-autonomy interface using Oculus Quest VR controllers. Our design uses a ``clutch'' mechanism that unifies policy execution and human takeover: when the side button is pressed, controller motions are mapped directly to the end effector enabling the human to take over control and intervene, and when the side button is released, the robot follows the learned policy. 
To reduce operator effort, we adopt a local-frame registration scheme with relative pose deltas. Let $v$ denote the fixed VR headset coordinate frame, and let $c_t$ denote the hand-controller frame at time $t$. At clutch engagement ($t=0$), we define the controller’s pose relative to the headset frame, $T^v_{c_0}$, as the local base frame. Subsequent poses are then expressed in this local frame as
$
T^{c_0}_{c}(t) = (T^v_{c_0})^{-1} T^v_{c_t},
$
with translational $\Delta p_k = p_k - p_{k-1}$ and rotational $\Delta R_k = R_{k-1}^\top R_k$ offsets used to parameterize end-effector commands. This design eliminates the need for global posture alignment, allowing operators to take over and intervene with minimal friction. A picture is shown in {Figure~\ref{fig:vr_handset}}.

\begin{wrapfigure}{r}{0.55\textwidth}
  \centering
  \vspace{-0.3cm}
  \includegraphics[width=0.99\linewidth]{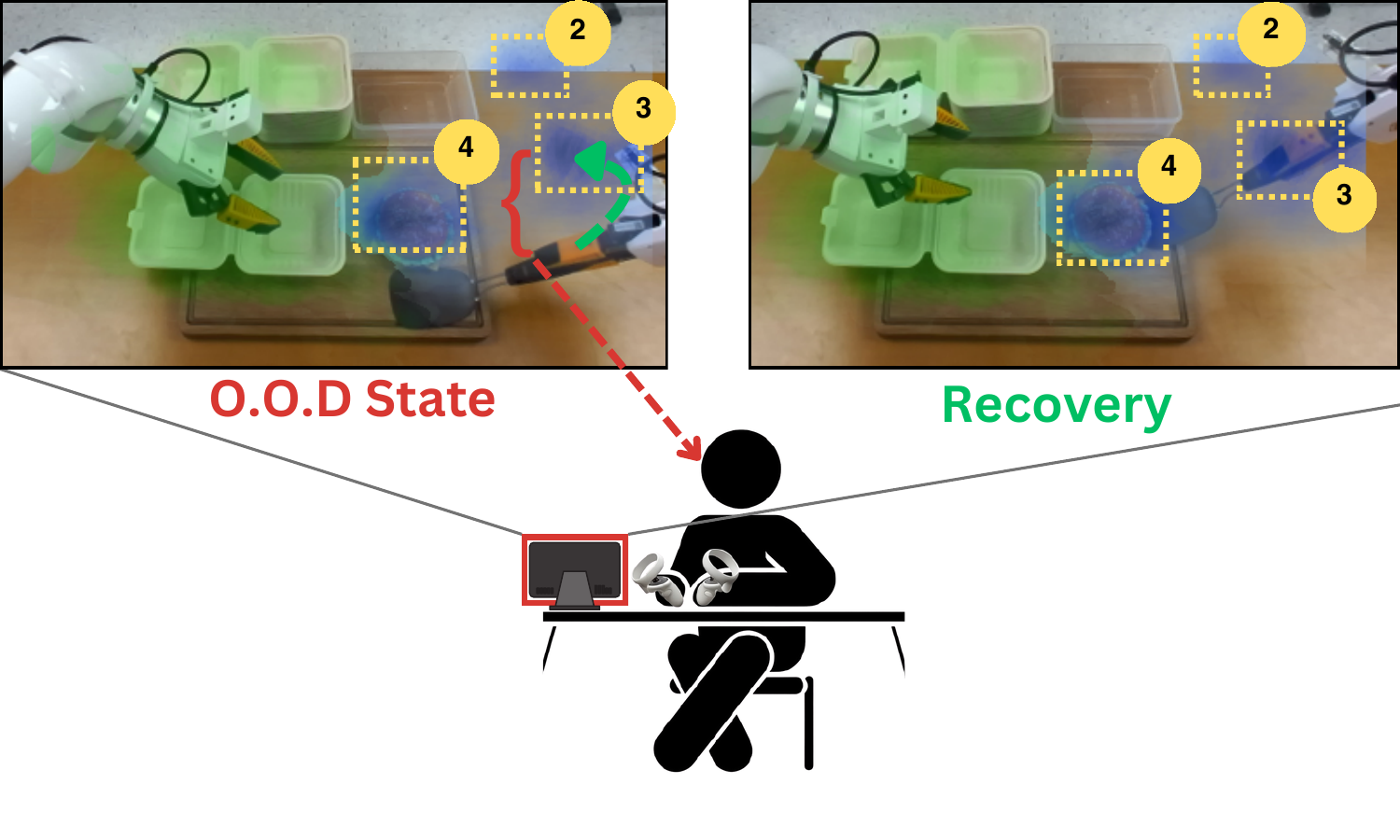}
  \vspace{-0.5cm}
  \caption{\footnotesize{\textbf{\emph{Visual aid for guiding intervention data collection.}} We utilize overlaid heatmap of the grippers visitation frequency to illustrate in-distribution regions that a teleoperator should recover to. In the \texttt{clamshell-takeout-box-packing} task, when the policy fails to scoop the burger using the spatula in sub-task 3 as shown above, the teleoperator recovers to the position inside the bounding box marked with sub-task 3 for retrying again.}}
  \label{fig:recover_tool}
  \vspace{-0.4cm}
\end{wrapfigure}
\emph{\textbf{Guidance for intervention data collection.}} To facilitate operators in demonstrating trajectories that adhere to the recovery then correction rule, we build a lightweight software tool using the image segmentation model SAM2~\citep{ravi2024sam2segmentimages}. This tool renders a robot end effector visitation frequency heatmap by tracking robot grippers across all RGB frames recorded by the overhead camera in the initial round of full demonstration data collection.
As shown in Figure \ref{fig:recover_tool}, during data collection, we overlay this heatmap onto the overhead camera's display window to provide visual aid, showing in-distribution regions where the robot grippers should recover back to upon intervention.
Our approach is one way that can be used to guide recovery demonstrations towards in-distribution regions.

\vspace{-0.35cm}
\subsection{Policy Architecture and Training via Imitation Learning}
\label{subsec:policy_arch}
\vspace{-0.15cm}

We now run imitation learning from a dataset containing multi-modal, long-horizon behaviors of various 
\begin{wrapfigure}{r}{0.5\textwidth}
  \centering
  % \vspace{-0.1cm}
  \includegraphics[width=\linewidth, clip,trim=0cm 1cm 0cm 2cm]{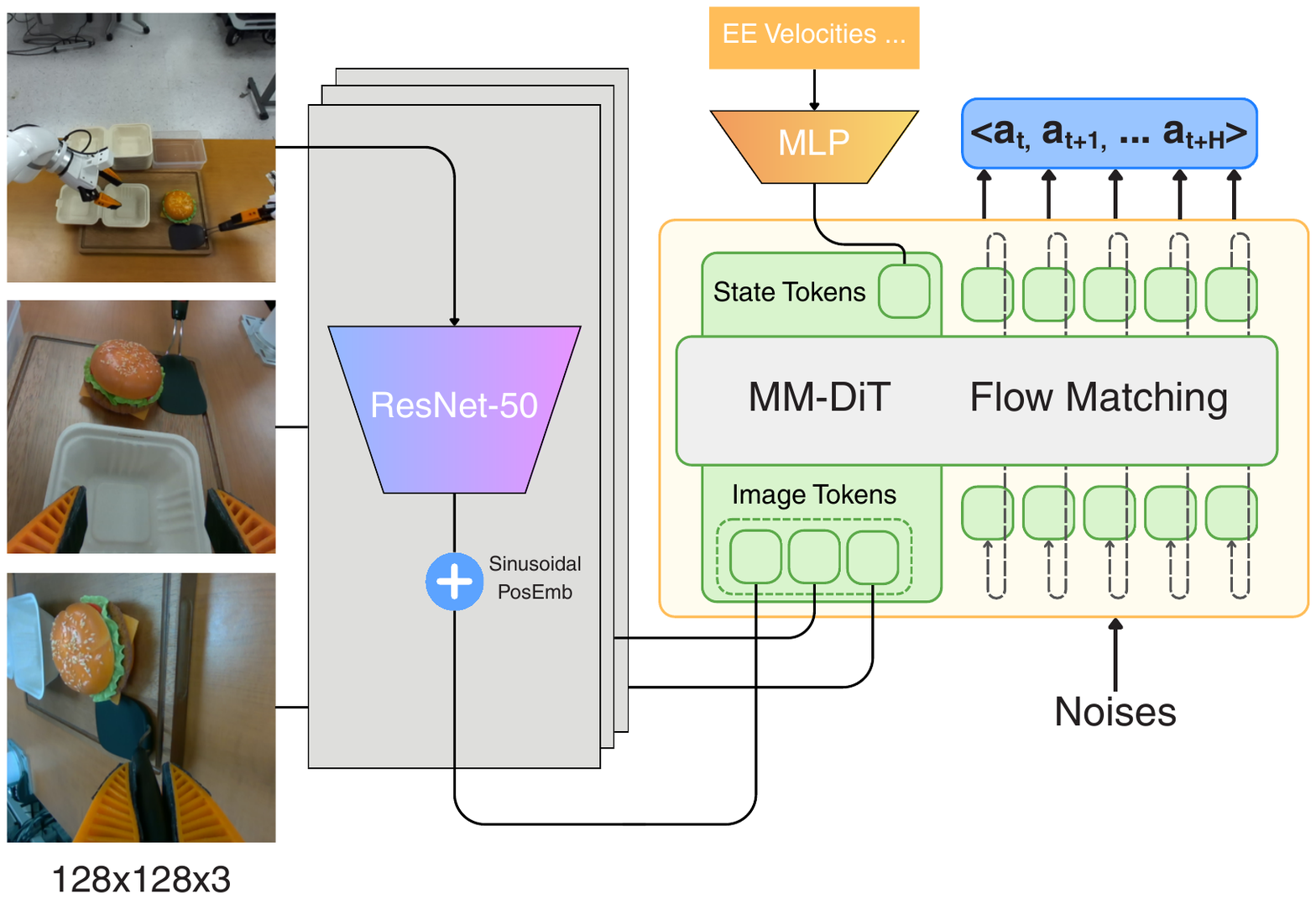}
  \vspace{-0.7cm}
  \caption{\footnotesize{\textbf{\emph{Policy architecture.}} We train all imitation learning policies using a multi-modal diffusion transformer (mm-DiT) architecture\citep{esser2024scalingrectifiedflowtransformers} via a flow matching objective.}}
  \label{fig:shirt_baselines}
  \vspace{-0.5cm}
\end{wrapfigure}
\!types: \textbf{1)} full demonstrations, \textbf{2)} the policy's own full successes from online rollouts, and \textbf{3)} human intervention segments with recoveries and corrections. Fitting various sources of data demands a high-capacity policy architecture, with sufficiently expressive output heads~\citep{chi2024diffusionpolicy, black2024pi0visionlanguageactionflowmodel, liu2024rdt}. Therefore, we utilize a flow-matching~\citep{lipman2022flow} policy to fit an action chunk~\citep{zhao2023learning}, $A_t = [a_t, a_{t+1}, ..., a_{t+H-1}]$ conditioned on observation $o_t= [I^1_t, I^2_t, I^3_t, q_t]$, where $I^i_t$ is the i-th RGB camera image and $q_t$ is a vector of robot states containing end effectors velocities and relative distance from each other at timestep $t$. For all tasks, we use $H=60$, equivalent to predicting one second of actions into the future.

Our policy is a 300 million parameter, multimodal diffusion transformer \texttt{(MM-DiT)} architecture~\citep{esser2024scalingrectifiedflowtransformers}. We use separate ResNet-50~\citep{he2016deep} vision encoders for all 3 camera views (1 overhead and 2 wrist cameras) in our real-world experiments and ResNet-18 encoders in simulation. We optimize a conditional flow matching loss~\citep{lipman2022flow} for training:
\begin{align}
\label{eq:flow_match}
\mathcal{L}_{\text{Flow}}(\theta) =
\mathbb{E}_{\substack{
o_t,\,A_t \sim \mathcal{D}, \\
x^0 \sim \mathcal{N}(0, I_d), \\
\tau \sim \text{Unif}([0,1])
}}
\left[ \left\| v_\theta(\tau, o_t, x^\tau) - \big(A_t - x^0\big) \right\|_2^2 \right],
\end{align}
where $x^\tau$ denotes an interpolant computed at time $\tau$ of the flow, $v_\theta(\tau, o_t, x^\tau): [0,1] \times S \times \mathbb{R}^d \rightarrow \mathbb{R}^d$ is velocity at $x^\tau$, and $d$ is the total dimensionality of action chunks we use. Importantly, when sampling training data from $\mathcal{D}$, we do not include any transitions from the robot's own rollouts, unless the trajectory reaches full task completion without any human intervention. This design choice is consistent with HG-DAgger~\citep{kelly2019hg, wu2025robocopilothumanintheloopinteractiveimitation}, but different from other methods such as IWR and follow-ups~\citep{mandlekar2020human, liu2022robot}, that filter segments based on human knowledge.
Additional details of policy training are in Appendix \ref{appendix:model_train_details}.

During inference, we generate actions by taking 10 Euler integration steps using the learned vector field from $t=0$ to $t=1$, starting with random noise $A^0_t \sim \mathcal{N}(0, I_d)$. Following~\citep{chi2024diffusionpolicy, chi2024universal, black2024pi0visionlanguageactionflowmodel}, we run policy inference once every 0.5 seconds, i.e., we execute the first half of each action chunk and then replan. A complete pseudocode of the procedure is shown in Algorithm~\ref{alg:methodname}.

% \vspace{-0.5cm}
% \subsection{Putting it All Together}
% \vspace{-0.25cm}
% In this section, we describe our complete approach, termed as \methodname{}, for practitioners.
% \vspace{-0.25cm}

\begin{algorithm}[t]
\caption{\methodname{}  Data Collection Protocol}
\label{alg:methodname}
\small
\begin{algorithmic}[1]
\footnotesize
\State Given per-round human data collection budget $\mathcal{B}$ measured in the number of frames; total human intervention data collection rounds $K$.
\State Initialize flow-matching policy $\pi_{\theta}^{k=0}$; dataset $\mathcal{D}_{0:K} \gets \varnothing$
\State Collect $\mathcal{B}$ frames of expert demonstrations in $ \Delta \mathcal D_0$; $\mathcal{D}_{0:K} \gets \Delta \mathcal D_0$; $\pi_{\theta}^{k=0} \gets \textsc{Train}(\mathcal{D}_{0:K})$ via Equation~\ref{eq:flow_match};
\end{algorithmic}

% \vspace{-0.1cm}
\textbf{Human Intervention Data Collection Rounds}
\begin{algorithmic}[1]
\footnotesize
\For{$k=1$ to $K$}
\State initialize human policy $\pi_H$, intervention function $I$
\State $\Delta \mathcal D_k \gets \varnothing$; \; $b \gets 0$ \Comment{$b$ data collection budget used in this round}
\While{$b < \mathcal{B}_k$}
  \State $s_0 \gets \texttt{env.reset()}; \; \texttt{traj} \gets [\,]; \; \texttt{intervened} \gets \textbf{false}; \; t \gets 0$
  \While{\textbf{not} $\texttt{env.done()}$}
    \If{$I(s_t)=0$} $a_t \sim \pi_{\theta}^{k-1}(\cdot \mid s_t)$; \; $\texttt{is\_human}\gets 0$
    \Else \; $a_t \sim \pi_H(\cdot \mid s_t)$; \; $\texttt{is\_human}\gets 1$; \; $\texttt{intervened}\gets \textbf{true}$ \Comment{Rule 1: Pair each recovery a correction}
    \EndIf
    \State $s_{t+1} \gets \texttt{env.step}(a_t)$; \; $\texttt{traj.push}(s_t, a_t, \texttt{is\_human})$; \; $t{+}{=}1$
    \If{$\texttt{is\_human}=0$ \textbf{and} $\textsc{InterventionDone}()$} \textbf{break} \Comment{Rule 2: Terminate after intervention concludes}
    \EndIf 
  \EndWhile
  \If{\texttt{intervened}=\textbf{false}} \Comment{If an entire trajectory has no human intervention $\Rightarrow$ }
    \State $\Delta \mathcal D_k \cup\!= \texttt{traj}$ \Comment{add full trajectory into dataset, with no human budget counted}
    % \State $b \gets b$ \Comment{}
  \Else
    \State $\Delta \mathcal D_k \cup\!= \{(s,a) \in \texttt{traj} : \texttt{is\_human}=1\}$ \Comment{add only human intervention transitions into dataset}
    \State $b = b + |\texttt{traj}|$ \Comment{charge full episode length to budget}
  \EndIf
\EndWhile
\State $\mathcal D_{0:K} \cup\!= \Delta \mathcal D_k$; \; $\pi_{\theta}^{k} \gets \textsc{Train}(\mathcal D_{0:K})$ \Comment{Aggregate datasets, then train policy via flow-matching \ref{eq:flow_match}}
\EndFor
\end{algorithmic}
\end{algorithm}

\vspace{-0.25cm}
\section{Experimental Evaluation of \methodname{}}
\label{sec:experiments}
\vspace{-0.25cm}
Our goal is to evaluate the data efficiency and scaling of \methodname{} on bimanual, long-horizon manipulation tasks. Concretely, we aim to answer the following questions: \textbf{(1)} Does \methodname{} improve data scaling compared to standard human full demonstration data collection, including existing state-of-the-art results?, \textbf{(2)} How does \methodname{} compare to human-in-the-loop imitation learning methods such as HG-DAgger~\citep{kelly2019hg}?, \textbf{(3)} Is enhancing the proportion of recovery behaviors critical for effective performance?, and \textbf{(4)} How do policies learned by \methodname{} differ from traditional imitation learning policies? We answer these questions through experiments in three real-world long horizon tasks. We also use a combination of real and simulated experiments to provide ablations to establish the role of recovery behaviors in training enabling test-time scaling on long-horizon tasks, with extra ablations on design choices of \methodname{} in Section~\ref{subsec:ablations}.

\vspace{-0.25cm}
\subsection{Evaluation Domains and Task Setups}
\label{subsec:task_descriptions}

\begin{wrapfigure}{r}{0.4\textwidth}
  \centering
  \vspace{0.05cm}
  \includegraphics[width=0.99\linewidth]{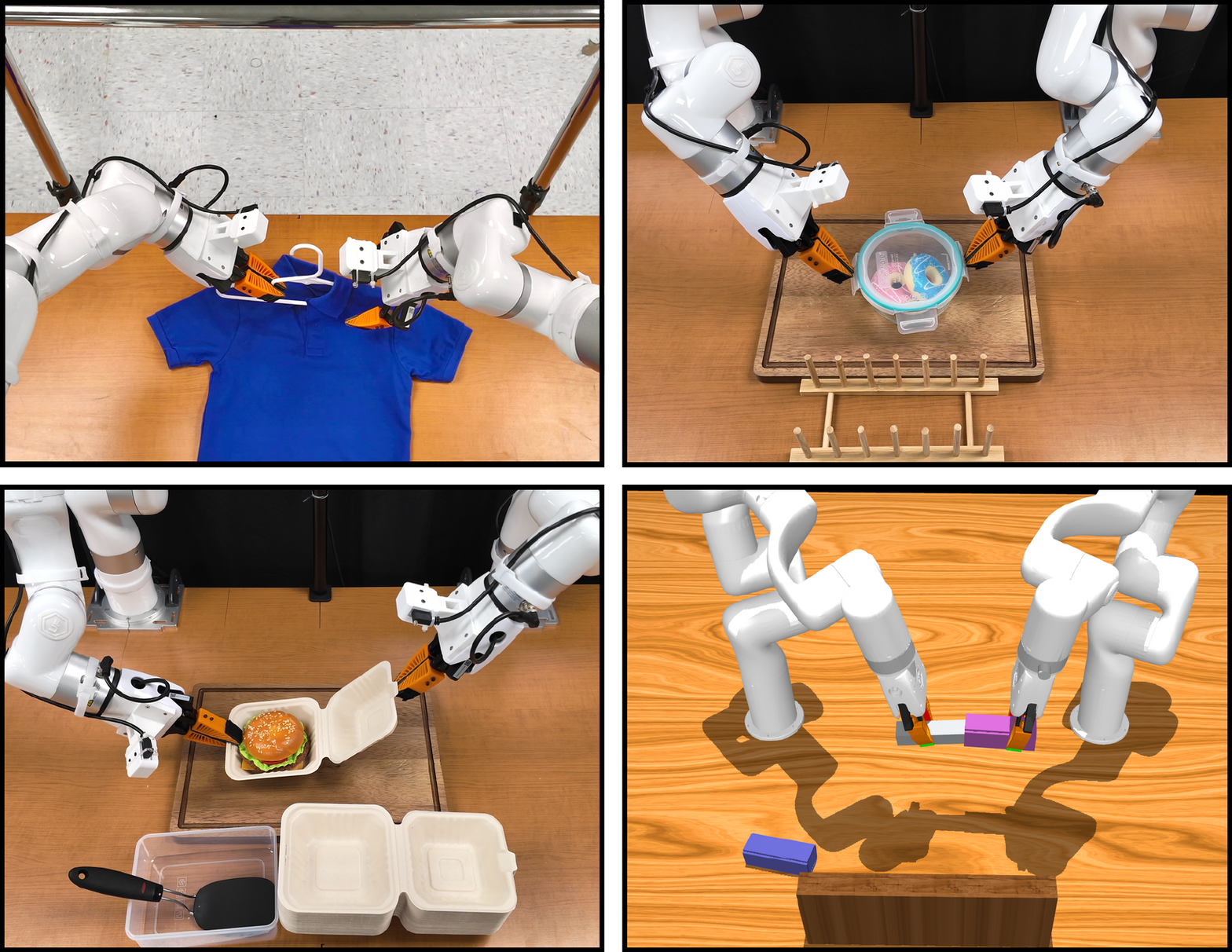}
  \vspace{-0.6cm}
  \caption{\footnotesize{\textbf{\emph{Long-horizon robot tasks.}} We study 3 real-world tasks, {shirt-hanging}, {airtight-container-lid-sealing}, {clamshell-takeout-box-packing}, and a simulated {bimanual-assembly} task.}}
  \label{fig:2x2_tasks}
  \vspace{-0.4cm}
\end{wrapfigure}

We design four bimanual dexterous manipulation tasks, with three in real world and one in simulation (see Figure \ref{fig:2x2_tasks}). All tasks require controlling the robot over extended horizons, demanding successful completion of interdependent sub-tasks sequentially. Our real-world tasks are inspired from some of the most difficult challenges explored in prior work~\citep{zhao2024alohaunleashedsimplerecipe,li2025tamingvrteleoperationlearning}. We briefly describe these tasks below and refer readers to detailed task definitions in the Appendix and show videos on our \href{https://rac-scaling-robot.github.io/}{website}.

\begin{itemize}[itemsep=4pt]
\vspace{-0.5cm}
    \item \texttt{shirt-hanging.} The robot \textbf{(1)} lifts a hanger off the rack; \textbf{(2)} hands over the hanger to the other gripper; \textbf{(3)} picks up and inserts the hanger into the right collar of the shirt; \textbf{(4)} picks up and inserts the hanger into the left collar; \textbf{(5)} hangs the shirt back on the clothing rack.
    \item \texttt{airtight-container-lid-sealing.} The robot \textbf{(1)} grasps a lid from a dish rack; \textbf{(2)} re-orients the gripper and places the lid on top of the container; \textbf{(3)} snaps two opposing tabs to partially seal; \textbf{(4)} rotates the bowl; \textbf{(5)} snaps the last two opposing tabs to complete the seal.
    \item \texttt{clamshell-takeout-box-packing.} The robot \textbf{(1)} grasps \emph{exactly one} takeout box from the stack and places it near the burger; \textbf{(2)} picks up the spatula; \textbf{(3)} scoops up the burger; \textbf{(4)} places the burger into the box; \textbf{(5)} nudges the burger to adjust its placement in the box; \textbf{(6)} closes the lid; \textbf{(7)} secures the lid by tucking the locking tab into its slot.
    \item \texttt{bimanual-assembly.} The robot \textbf{(1)} picks up and inserts white block into blue socket; \textbf{(2)} picks up pink socket and inserts it on top of white block; \textbf{(3)} places the total workpiece on the platform.
    \vspace{-0.3cm}
\end{itemize}

\textbf{Comparisons and evaluation protocol.} We compare the scaling characteristics, performance, and learned behaviors of \methodname{} against two approaches for imitation learning: \textbf{(1)} scaling up batched full expert
\!\!data collection, and \textbf{(2)} performing human-in-the-loop interventions as per HG-DAgger~\citep{kelly2019hg}. For each task, we allocate a total budget of $K\!\!\times\!\!N$ demonstrations for the batched setting, where $N$ is a base number of demonstrations chosen in advance. To match this budget, we run $K$ rounds of human-in-the-loop data collection, each with equivalent per-round budget, and train the policy in each round using the corresponding intervention data. We conduct evaluations with 60 trials for the real-world tasks and 100 trials in the simulation task with varying initial configurations (details in Appendix \ref{appendix:task_eval_protocols} and videos on \href{https://rac-scaling-robot.github.io/}{website}). When rolling the trained policy out during evaluation, we record the performance for each sub-task upto an irrecoverable failure, then we terminate the episode. We measure sub-task performance per a binary success or failure indicator function without assigning partial credits. We report two performance metrics: task success rates and task progress scores. Task success rates indicate the percentage of trials that completes all sub-tasks, while task progress scores represent the number of sub-tasks a trial is able to complete.

\vspace{-0.3cm}
\subsection{Main Results: One Order of Magnitude Improvement in Data Efficiency}
\label{subsec:main_results_data_efficiency}
\vspace{-0.2cm}
\begin{figure}[t]
\vspace{-0.5cm}
  \centering
  \includegraphics[width=0.99\linewidth]{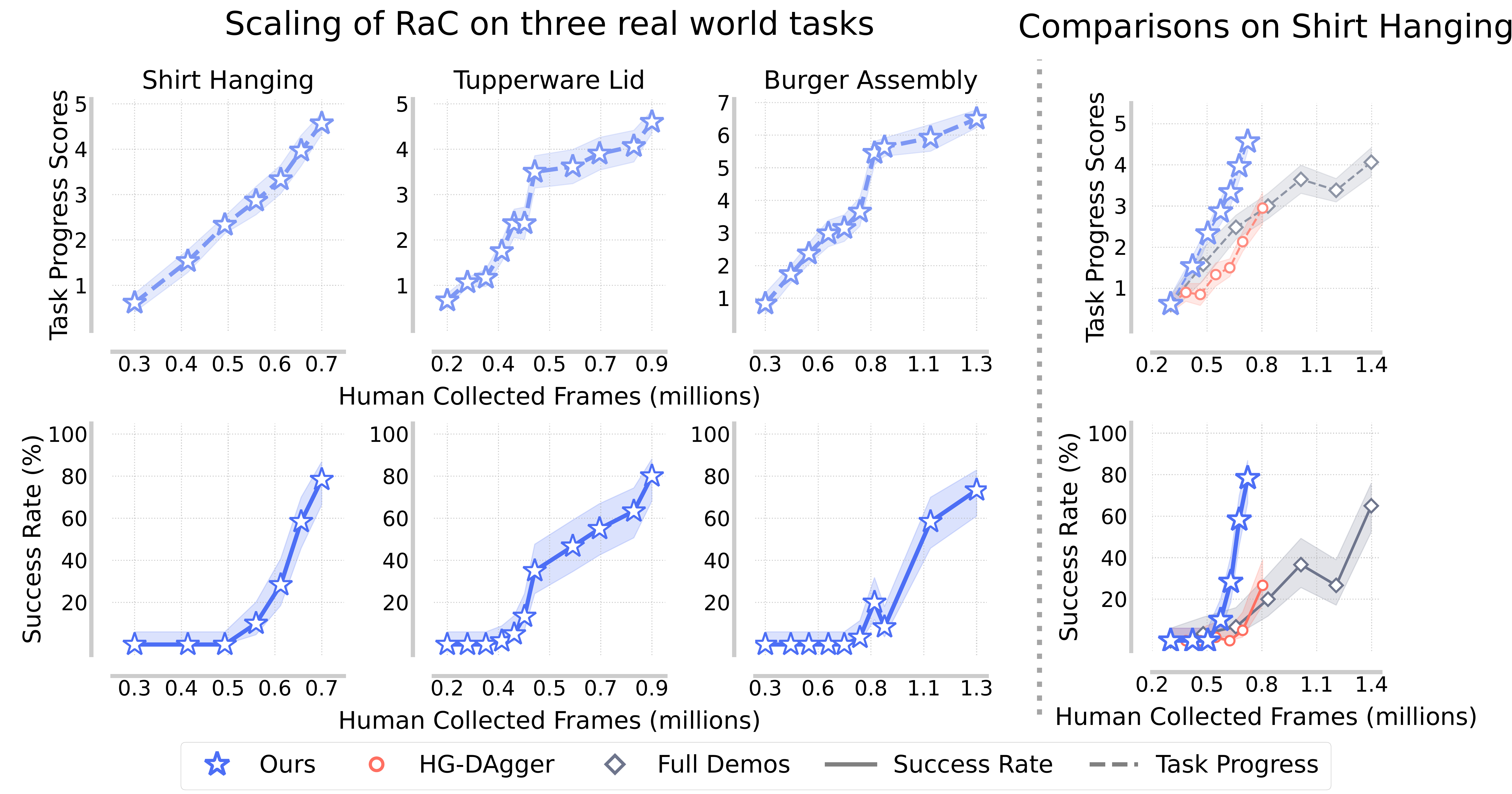}
  \vspace{-0.25cm}
  \caption{\footnotesize{\textbf{\emph{Performance scaling for \methodname{}}} as a function of human-collected frames on real-world tasks. Note that within $K=6$ rounds for {shirt-hanging}, $K=10$ rounds for {airtight-lid-sealing}, and $K=9$ rounds for {takeout-box-packing}, we observe the best-known results for tasks of a similar difficulty from prior work. The top row shows average progress over various sub-tasks,  the bottom row shows full long-horizon task success rate. On the right, we compare \methodname{} to various other baseline approaches based on HG-DAgger and cloning full demonstration data, and observe a substantial improvement in data efficiency.}}
  \vspace{-0.5cm}
  \label{fig:performance_2x3}
\end{figure}
Despite the challenges associated with coherent long-horizon execution, deformable object handling, and contact-rich manipulation, our policies reach high success rates and task progress scores with only modest data requirements. Strikingly, just 5 hours of training data suffice to surpass the full success rates of 75\%. To highlight data efficiency gains, consider the \emph{shirt-hanging} task: prior works~\citep{zhao2024alohaunleashedsimplerecipe, cheang2025gr3technicalreport} report needing thousands of expert demonstrations or more than one hundred hours of teleoperation data to achieve a comparable success rate to \methodname{}. \methodname{} achieves similar or better results with an \textbf{\emph{order of magnitude}} less data, illustrating its efficacy in scaling imitation learning (Table~\ref{tab:prior}  and Figure \ref{fig:performance_2x3}).

% Table
\begin{table}[t]
  \centering
  \footnotesize
  \setlength{\tabcolsep}{4pt}    % adjust column padding
  \renewcommand{\arraystretch}{1.25}
  \begin{tabularx}{\linewidth}{l c c Y r}
    \toprule
    \textbf{Name} & \textbf{Policy Architecture} & \textbf{Model Size} & \textbf{Training Data Size} & \textbf{SR} \\
    \midrule
    ALOHA Unleashed~\citep{zhao2024alohaunleashedsimplerecipe} & Diffusion Transformer policy & 217M & $\sim\!89$ hours (5345 shirt-hanging expert demos) & $75.0\%$ \\
    Seed GR-3~\citep{cheang2025gr3technicalreport}       & Vision-Language-Action model & 4B & 116 hours of shirt-hanging expert demos and vision-language data & $\sim\!63.6\%$ \\
    Ours (\methodname{}) & Flow-matching Transformer policy & 368M & \textbf{5 hours} (\methodname{} data: expert, recovery, and correction) & \textbf{78.3\%} \\
    \bottomrule
  \end{tabularx}
  \vspace{-0.2cm}
  \caption{\footnotesize{\emph{\textbf{Comparison to similar shirt-hanging tasks in prior work.}} Under similar task setups and difficulty, the full task success rate (``SR'' of \methodname{} policy is higher than other methods using an order of magnitude less data. See Appendix \ref{appendix:shirt_comparison} for details.}}
  \label{tab:prior}
\end{table}

\textbf{Comparisons on real-world tasks.} Since scaling up batched data collection across all real-world tasks was infeasible due to the prohibitive costs of collecting such large expert datasets, we instead scaled the batched data collection baseline on one representative task, {shirt-hanging}. Observe in Figure~\ref{fig:shirt_baselines}, \methodname{} not only achieves substantially higher absolute performance and task progress, but also delivers at least a \textbf{2}$\times$ improvement in data efficiency compared to the batched data collection approach. \methodname{} also consistently outperforms HG-DAgger. This result does not arise from a subpar baseline: our HG-DAgger implementation exhibits performance trends consistent with prior work, 
such that it outperforms batched data collection under the same amount of human collected data. Finally, we note that \methodname{} exhibits a markedly steeper scaling curve (``higher slope'') than either baseline in Figure~\ref{fig:shirt_baselines}.

\vspace{-0.25cm}
\subsection{Examining the Properties of Policies Learned via \methodname{}}
\label{subsec:learn_dynamics}
\vspace{-0.15cm}

\begin{figure}[t]
  \centering
  \vspace{-0.1cm}
  \includegraphics[width=0.485\linewidth]{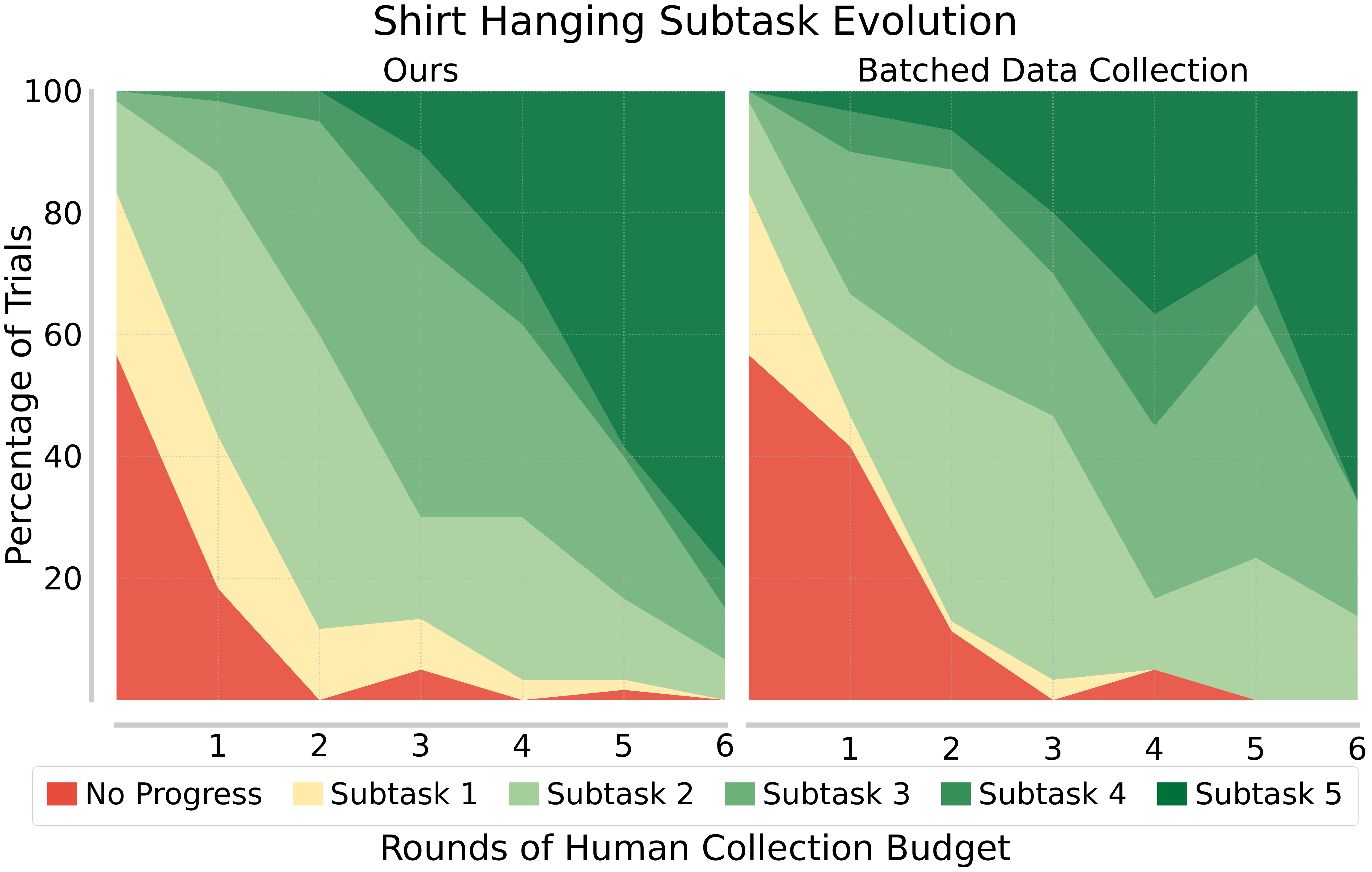}
  ~\vline~
  \includegraphics[width=0.485\linewidth]{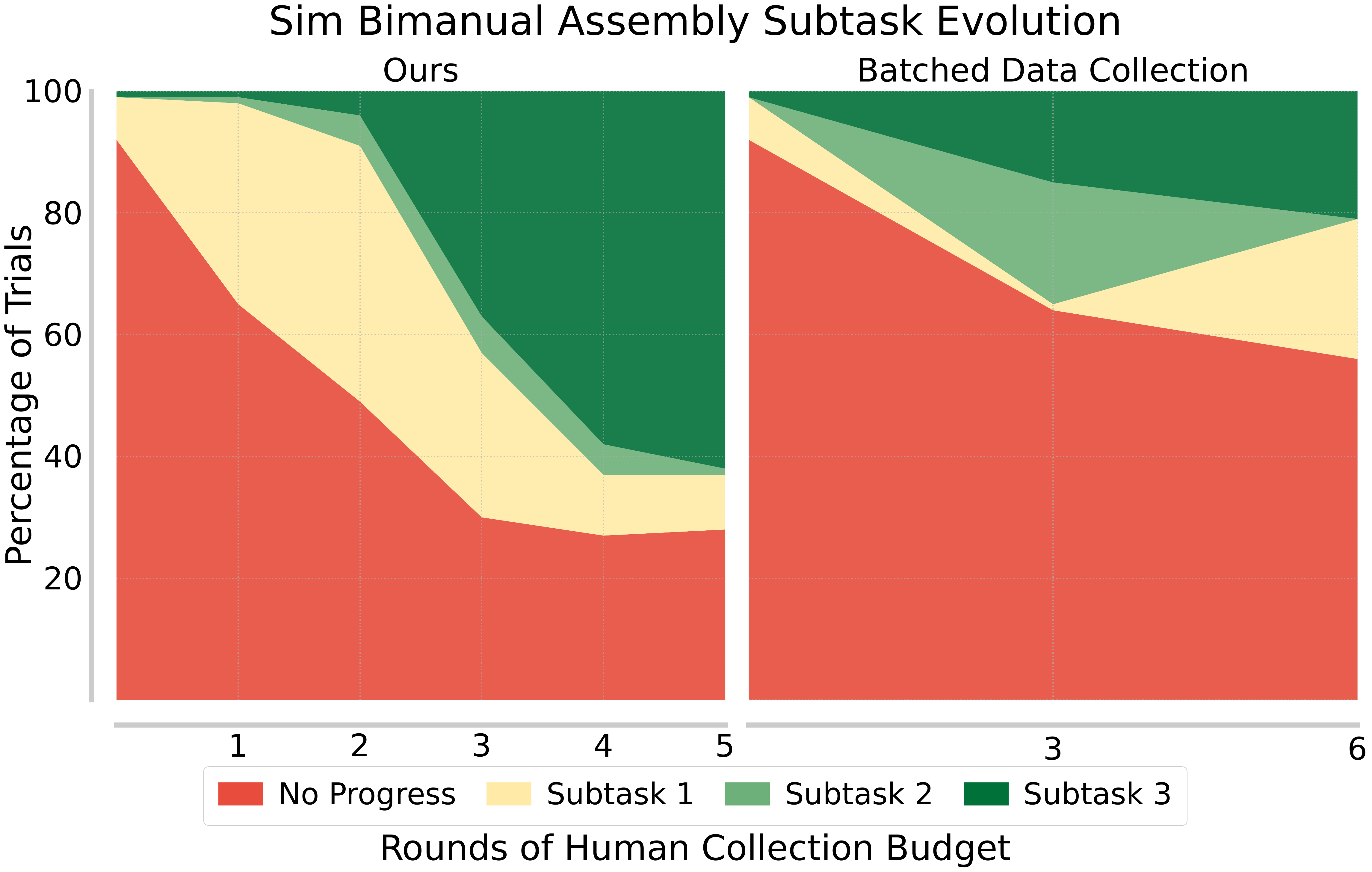}
  \vspace{-0.2cm}
  \caption{\footnotesize{\emph{\textbf{Performance profiles for \methodname{} and the batched data collection approach.}} For both real-world shirt-hanging (left two plots) and simulation assembly (right two plots), \methodname{} rapidly reduces the fraction of rollouts that make little progress and steadily shift probability mass toward later sub-task completions and full success. This trend however is not consistent or strong enough for various sub-tasks in the case of batched data collection, which trains on full demonstrations.}}
  \label{fig:watermelon}
  \vspace{-0.3cm}
  % \vspace{-1cm}
\end{figure}

\emph{\textcolor{lightblue}{\textbf{Result 1: Robustness of intermediate \methodname{} policies.}}} Having established the efficacy of \methodname{}, we next analyze the properties of the learned policies in a more systematic manner. To this end, we visualize in Figure~\ref{fig:watermelon}, the distribution of sub-tasks completed by intermediate policy checkpoints produced during successive rounds of human intervention (for \methodname{}) and as we scale data (for batched data collection). We observe that the fraction of on-policy rollouts making little progress rapidly decreases with more rounds when using \methodname{}. In other words, \methodname{} systematically reduces/eliminates the long tail of rollouts that fail or stall early. In contrast, training on increasing amounts of batched full demonstration data does not exhibit the same kind of progress on all sub-tasks, especially in simulation (Figure~\ref{fig:watermelon}, right).
Because our evaluations begin from a broad set of initial configurations, this experiment in a sense highlights the robustness of \methodname{}. To summarize, by explicitly scaling recovery, \methodname{} drives progress even in the difficult ``tail'' cases, a persistent failure mode that is common lore with imitation learning.

\begin{figure}[t]
% \vspace{0cm}
    \centering
    \vspace{-0.2cm}
    \includegraphics[width=0.8\linewidth]{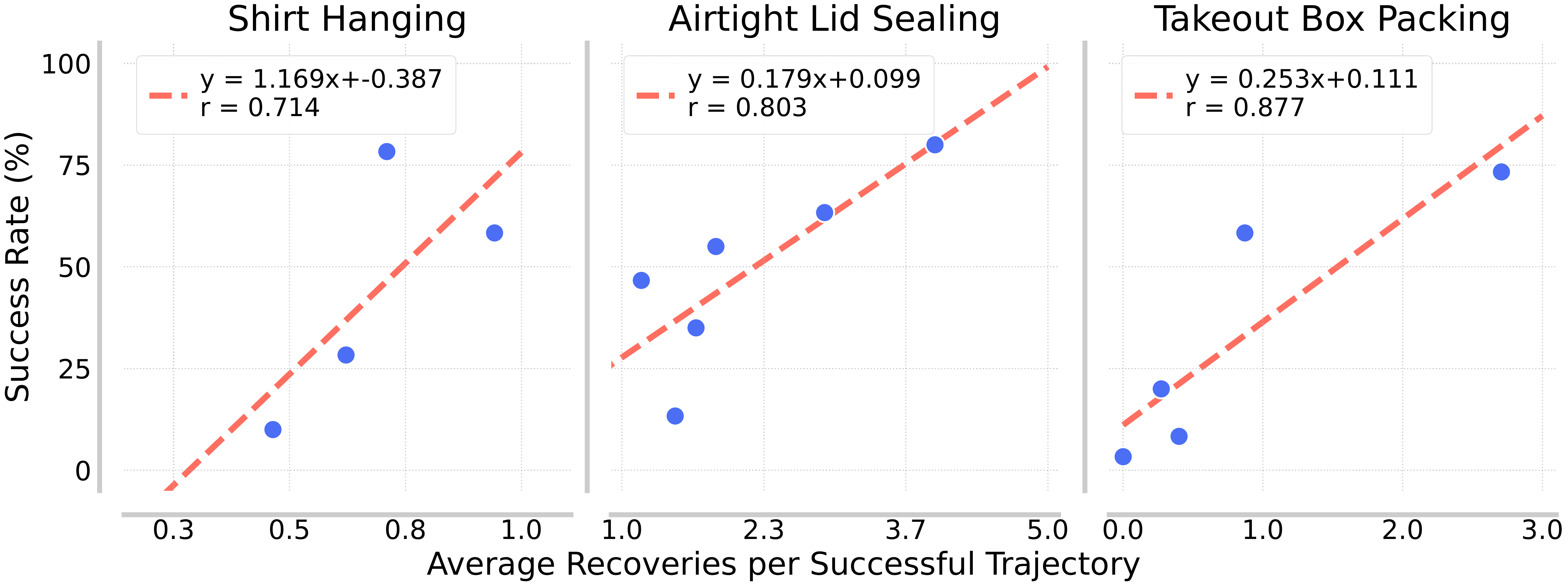}
    \vspace{-0.2cm}
    \caption{\footnotesize{\emph{\textbf{Test-time scaling of \methodname{} policies with the number of recovery segments.}} We observe a strong linear scaling relationship between the number of recovery segments upon policy deployment and success rate of policies produced by later rounds of \methodname{}. This is a form of test-time scaling analogous to that in LLMs~\citep{openai_learning_to_reason_2024}.}}
    \label{fig:lid_retries_vs_success}
    \vspace{-0.3cm}
\end{figure}
\textcolor{lightblue}{\emph{\textbf{Result 2: ``o1-style'' test-time scaling for robotic policies.}}} Next, we study whether performance scales with more recovery behavior at deployment. To do so, we analyze the subset of evaluation rollouts that successfully solve all sub-tasks across different rounds, and annotate each rollout with the number of recovery attempts it contains. In Figure~\ref{fig:lid_retries_vs_success}, we show the average number of recovery segments observed against the task success rates. The correlation coefficients $r$ indicate a linear relationship between the
\begin{wrapfigure}{r}{0.55\linewidth}
% \vspace{0cm}
    \centering
    \vspace{-0.4cm}
    \includegraphics[width=0.97\linewidth]{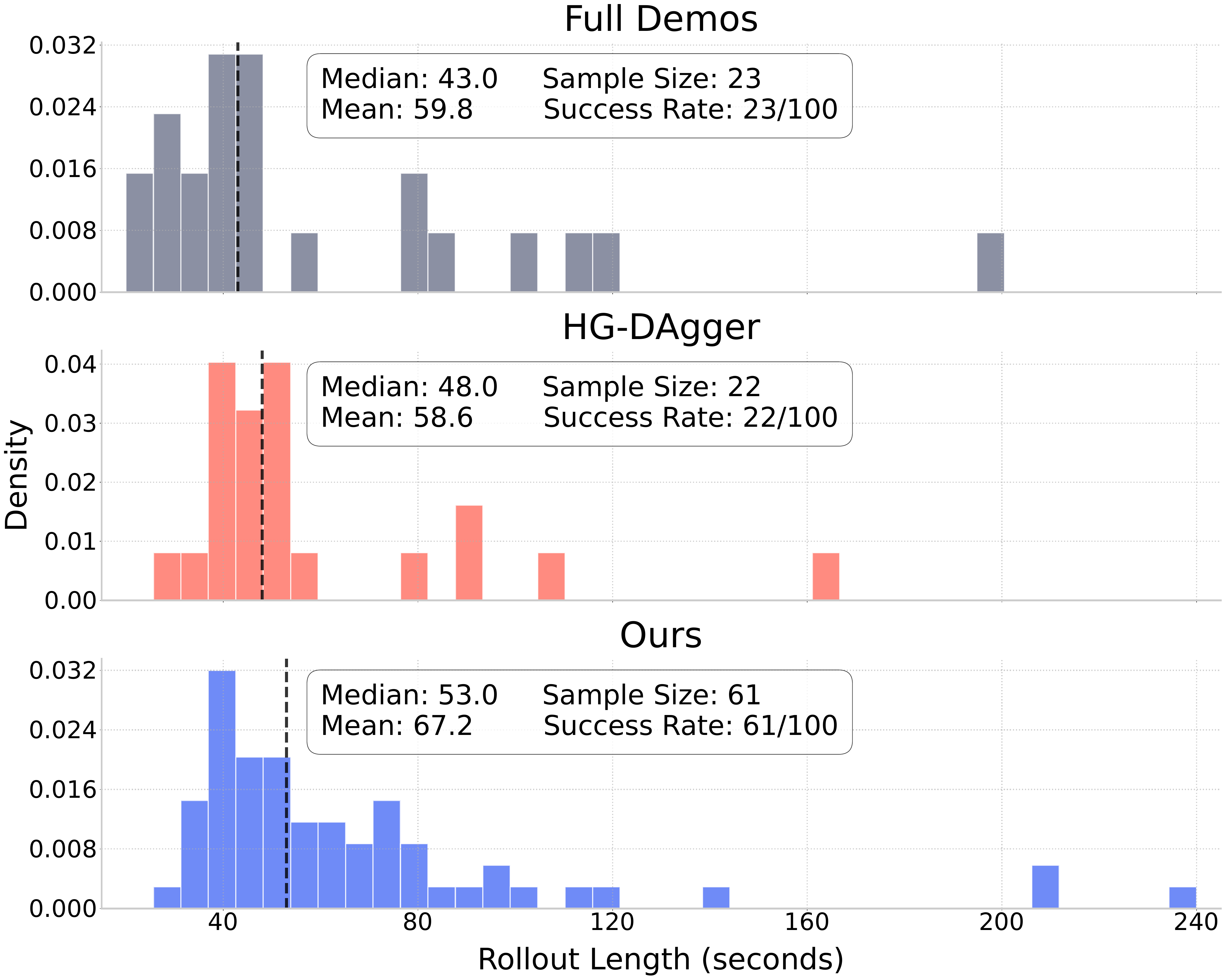}
    \vspace{-0.2cm}
    \caption{\footnotesize{\emph{\textbf{Distribution of lengths of successful rollouts}} for various approaches. Note that \methodname{} policies produce the longest rollouts on average due to the presence of recovery behavior. HG-DAgger produces the second-highest median rollout length.}}
    \label{fig:rollout_length_histogram}
    \vspace{-0.75cm}
\end{wrapfigure}
task success and recovery frequency. In other words, as the policy learns to demonstrate more recovery behaviors, its overall performance improves. To those readers familiar with LLMs, this pattern resembles 
favorable test-time scaling~\citep{openai_learning_to_reason_2024}: just as reasoning LLMs~\citep{openai_learning_to_reason_2024,guo2025deepseek} perform better when they produce longer CoTs that illustrate backtracking and error correction, robot policies that scale the number of recovery segments directly in the space of action sequences are likely to succeed more.

\textcolor{lightblue}{\textbf{\emph{Result 3: Rollouts from \methodname{} policies are generally longer, and more successful.}}} On the simulation task, we analyzed the wall-clock duration of successful evaluation rollouts across methods (Figure~\ref{fig:rollout_length_histogram}). 
Successful rollouts from \methodname{} are skewed towards longer lengths, reflecting recovery behaviors that keep the task on track. For \methodname{}, longer length is also correlated with better average performance and more successful rollouts. Successful HG-DAgger rollouts attain the second highest median length, since the robot is trained to still utilize corrective segments to succeed from out-of-distribution states. Policies trained on full demonstration data can likely only succeed when they stay within distribution, resulting in shortest median successful rollout length. Though interestingly, this approach also produces one outlier rollout that attains longer lengths compared to HG-DAgger, where the robot arms kept applying excessive force until the insertion succeeds, without explicitly recovering or correcting from failure states.

\vspace{-0.25cm}
\subsection{Ablation Studies for the \methodname{} Data Collection Protocol}
\label{subsec:ablations}
\vspace{-0.15cm}
Finally, we present ablation studies to better understand the properties of the human intervention data collected by \methodname{} across training rounds. In Figure \ref{fig:lid_retries_vs_success}, we visualize the composition of intervention data over 4 rounds in the simulation task. We compare data collected using the full \methodname{} approach (``Ours'') and \methodname{} without enforcing `recover-then-correct' (``Ours w/o Rule 1'', i.e. HG-DAgger with only Rule 2). Recall that these Rules were prescribed in Section~\ref{subsec:rac_rules}. 

\begin{figure}[t]
  \centering
  % \vspace{-0.3cm}
  \includegraphics[width=0.7\linewidth]{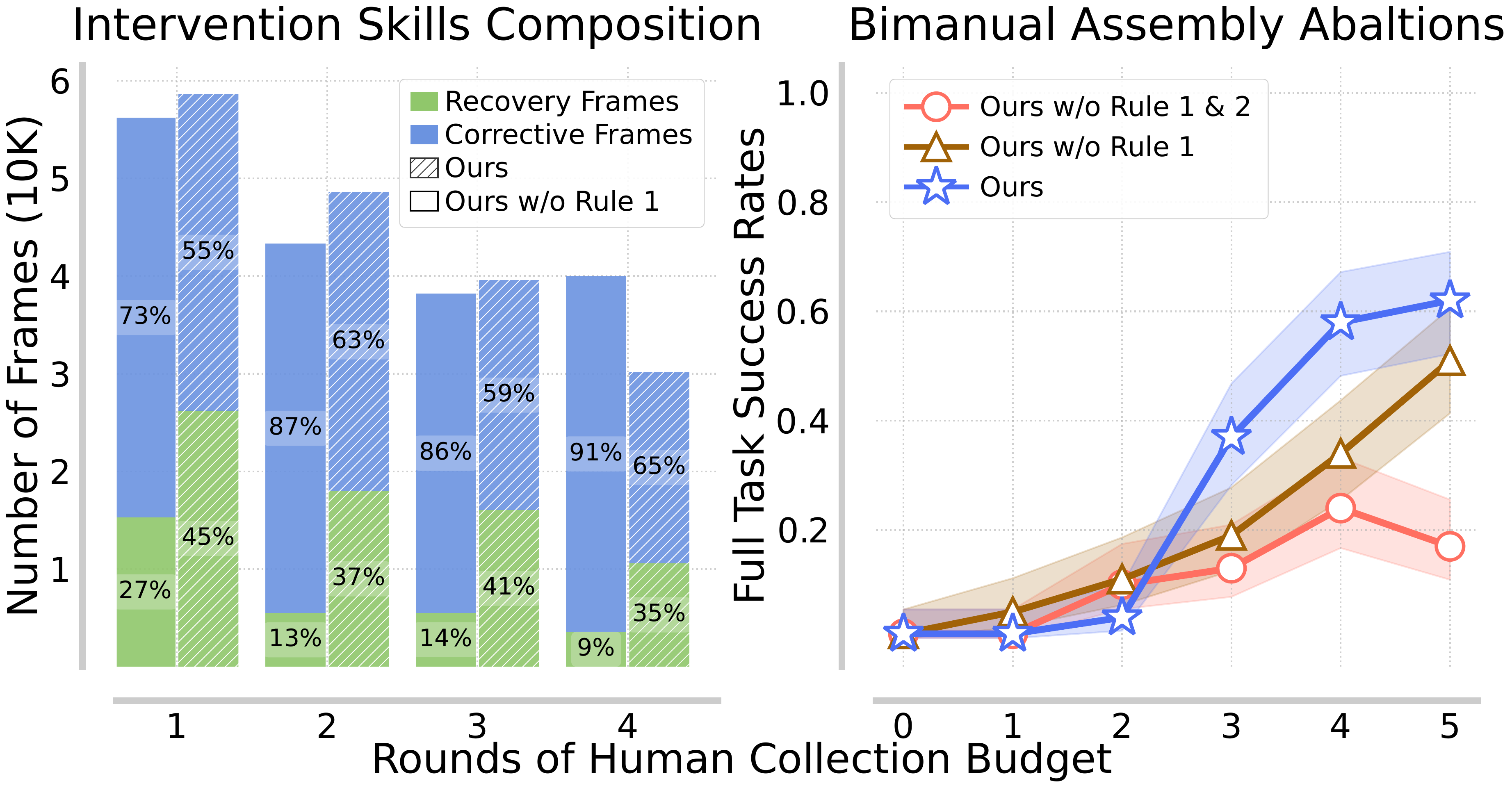}
  % \vspace{-0.2cm}
  \caption{\footnotesize{\textbf{\emph{Ablation studies on the simulation task.}} \textbf{Left:} Assessing the composition of human intervention data collected in each round. Note that data collected via \methodname{} maintains a high proportion of recovery segments along with corrective frames. On the other hand, the intervention data collected by HG-DAgger skews heavily towards corrective frames. \textbf{Right:} Utilizing ``Rule 2'' and terminating the intervention episode early yields better data scaling of performance than continuing the policy rollout after the recover-then-correct intervention is complete. This showcases the importance of both rules prescribed by \methodname{}.}}
  % \vspace{-0.3cm}
  \label{fig:sim_ablation}
\end{figure}

We classify each intervention frame as either a recovery segment or a corrective segment. Observe in Figure~\ref{fig:sim_ablation} (left), that while \methodname{} maintains a roughly balanced ratio of recovery to corrective frames (between 1:1 and 1:2), conventional intervention data exhibits a highly skewed distribution dominated by corrective frames, with recovery frame's proportion decreasing sharply in later rounds. Specifically, conventional intervention data contains 1:3 and 1:10 proportions of recovery/correction data. The total number of intervention frames naturally decreases as policies improve and require fewer interventions.

Next, we study the effect of Rule 2 in \methodname{}: truncating an episode after human intervention concludes. In the simulation task (Figure \ref{fig:sim_ablation}), we observe that terminating early after an intervention alone (``Ours w/o Rule 1'') yields more effective performance scaling than continuing policy rollouts after human intervention (``Ours w/o Rule 1\&2'', i.e., HG-DAgger). We hypothesize that this effect arises because allowing the rollout to continue after human intervention completes contaminates later parts of the trajectory with states influenced by both the human and the policy, producing data that are out-of-distribution for a learned policy. By terminating right after the intervention, we ensure that the collected data cleanly reflect recovery–correction behavior, while subsequent sub-tasks are reached only with the policy’s own distribution in future rounds, leading to more efficient data scaling.

\vspace{-0.25cm}
\section{Discussion, Conclusion, and Future Work}
\label{sec:discussion}
\vspace{-0.25cm}
We presented an approach, \methodname{}, for scaling imitation learning in the real world. Our core idea is to scale not just the quantity of data, but the \emph{type} of data, explicitly pairing recovery and correction behaviors collected through human interventions. By doing so, we enabled policies to mitigate compounding errors, retry from failures, and achieve substantially higher data efficiency than standard teleoperation or correction-only approaches. Our experiments demonstrated that this paradigm yields robust policies on long-horizon, contact-rich tasks with orders of magnitude less data than prior work and much better data efficiency than our comparisons. We also illustrated a form of ``test-time scaling'' by showing that more recovery segments and longer action times correlate with higher performance. 

\textbf{Future work.} We believe that there are quite a few avenues for future work. First, analogous to how autonomous RL began performing substantially better on top of properly mid-trained initializations for LLMs~\citep{wang2025octothinkermidtrainingincentivizesreinforcement}, we believe that policies trained via \methodname{} bear the potential to serve as good initializations for online RL fine-tuning on a real robot. Unlike typical imitation pre-trained policies that attempt to perform ``optimal'' behavior (and typically lose track upon failing to accomplish the task), we hypothesize that policies from \methodname{} would naturally provide more structured exploration and coverage during online RL due to the presence of recovery behavior. Recovery provides natural ``stitching'' points~\citep{fu2020d4rl} which might also be amenable to value-based training. Another interesting direction for future work is to apply \methodname{} on top of generalist vision-language-action (VLA) models~\citep{black2024pi0visionlanguageactionflowmodel,kim24openvla,geminiroboticsteam2025geminiroboticsbringingai}. Finally, while prior results do show some examples of recovery behaviors in VLA models, it is unclear if such behaviors systematically emerge in most settings or not, and studying this aspect rigorously (for example, by plotting test-time scaling curves analogous to Figure~\ref{fig:lid_retries_vs_success}) is also useful for the community.

\vspace{-0.25cm}
\section*{Acknowledgements}
\vspace{-0.2cm}

We thank Yuxiao Qu, Bhavya Agrawalla, Lehong Wu, Max Sobol Mark, Anikait
Singh, Yufei Wang, Divyam Goel, and Yiran Tao for feedback on an earlier version of this paper. We thank Jason Jingzhou Liu for help with RMPFLow infrastructure. We thank the members of CMU AIRe and RCHI labs for support and feedback. AK thanks Abhishek Gupta, Dhruv Shah, Amrith Setlur, and Max Simchowitz for informative discussions and feedback. This work was supported in part by an Apple seed grant, the Office of Naval Research under N00014-24-12206, and National Institute of Biomedical Imaging and Bioengineering of the National Institutes of Health under award number 1R01EB036842-01. We thank the Babel compute cluster at CMU, the TRC program of Google Cloud, and the National Centre for Supercomputing Applications for providing computational resources that supported this work.

\newpage

\bibliography{main}

\newpage
\appendix
\onecolumn
\part*{Appendices}

\section{Policy Architecture and Training Details}
\label{appendix:model_train_details}

We train all \methodname{} imitation learning policies with the same model architecture and training configurations detailed below. With the multi-modal DiT (mm-dit) architecture \citep{esser2024scalingrectifiedflowtransformers}, we use two separate modalities, i.e. two sets of transformer weights to model action generation conditioned on robot observations. The first set of transformer weights processes robot observations, including image tokens from the three camera views after ResNet encoders and a robot proprioceptive state token after a MLP encoder. The second set of transformer weights processes noised action tokens. mm-DiT joins the sequences of the two modalities for the attention operation, such that both representations can work in their own spaces while taking the other one into account. This design is similar to the action expert in \citep{black2024pi0visionlanguageactionflowmodel}.

ResNet encoders used in this work finetune on weights pre-trained on ImageNet.

All model trainings are conducted on 4-cards of RTX 6000 Ada GPU servers or 8-cards of L40S GPU servers. 

% -----------------------------
% Training Config Table
% -----------------------------
\begin{table}[h]
\centering
\small
\caption{\textbf{Model training configurations.} Training hyperparameters used for all experiments.}
\begin{tabular}{lc}
\toprule
\textbf{Config} & \textbf{Value} \\
\midrule
Optimizer & AdamW (default) \\
Learning Rate & $1 \times 10^{-4}$ (const.) \\
Global Batch Size & 512 \\
Training Length & 200 epochs \\
State Dimension & 40 \\
Action Dimension & 14 \\
Action Horizon & 60 \\
\bottomrule
\end{tabular}
\end{table}

% -----------------------------
% Flow Matching Model Details
% -----------------------------
\begin{table}[h]
\centering
\small
\caption{\textbf{Flow-matching policy details.} Architecture specifications of our policy model.}
\begin{tabular}{lc}
\toprule
\textbf{Detail} & \textbf{Value} \\
\midrule
\texttt{MM-DiT} modalities~\citep{esser2024scalingrectifiedflowtransformers} & 2 \\
Flow Matching Steps & 10 \\
MM-DiT Hidden Size & 768 \\
MM-DiT Depth & 12 \\
MM-DiT Heads & 12 \\
Vision Encoder & ResNet-50 (real) / ResNet-18 (sim) \\
Total Parameters & 367.865M \\
\bottomrule
\end{tabular}
\end{table}

\newpage

\section{Example Rollouts on Various Tasks}
\vspace{-0.15cm}

\begin{figure}[h]
  \centering
  % \vspace{-0.15cm}
  \includegraphics[width=0.99\linewidth]{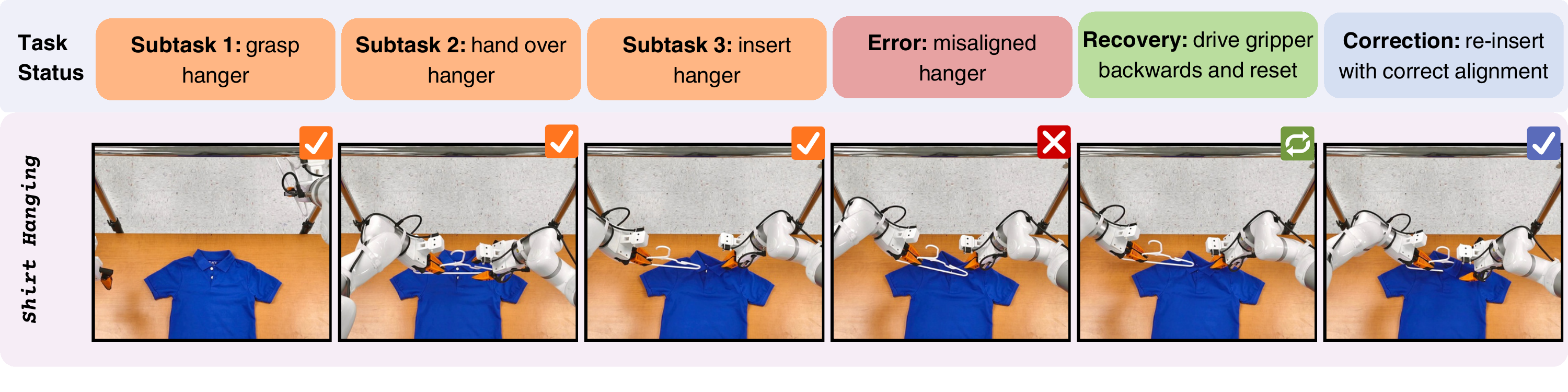}
  \vspace{-0.25cm}
  \caption{\footnotesize{\emph{\textbf{\methodname{} rollout on the shirt-hanging task.}} In this task, recovery corresponds to driving the gripper and hanger backwards and correction corresponds to reinserting the hanger again.}}
  \label{fig:shirt_filmstrip}
  \vspace{-0.3cm}
\end{figure}
% \vspace{-0.30cm}

\begin{figure}[h]
  \centering
  % \vspace{-0.15cm}
  \includegraphics[width=0.99\linewidth]{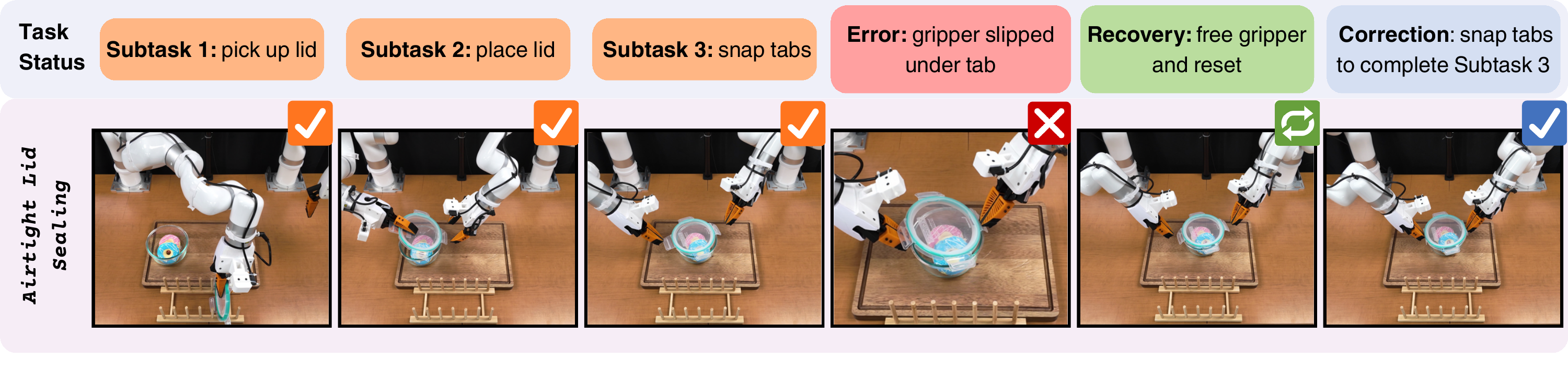}
  \vspace{-0.25cm}
  \caption{\footnotesize{\emph{\textbf{\methodname{} rollout on the airtight-container-lid-sealing task.}} In this task, recovery corresponds to driving the gripper and hanger backwards and correction corresponds to reinserting the hanger again.}}
  \label{fig:lid_filmstrip}
  \vspace{-0.3cm}
\end{figure}
% \vspace{-0.30cm}

\begin{figure}[h]
  \centering
  % \vspace{-0.15cm}
  \includegraphics[width=0.99\linewidth]{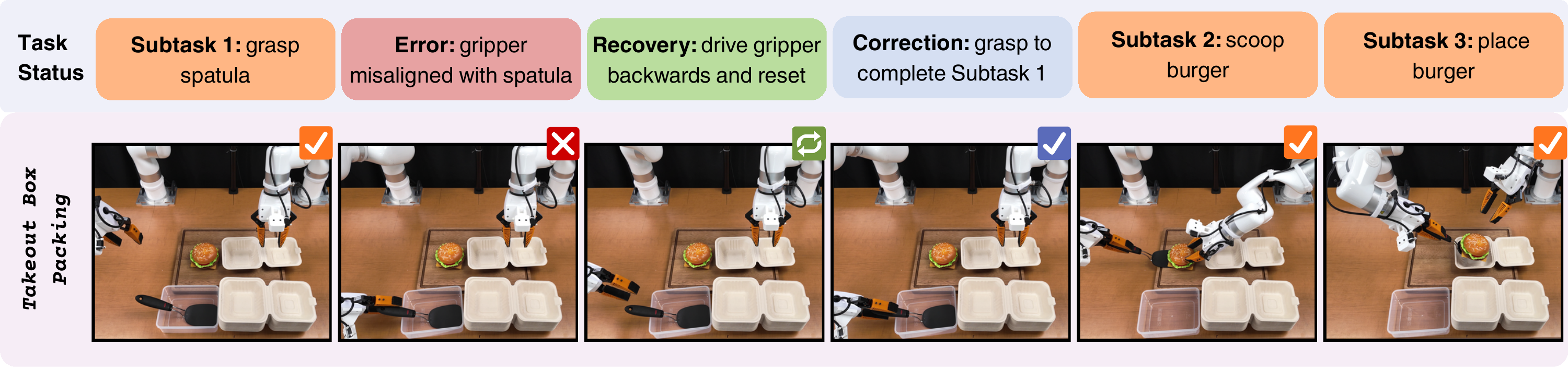}
  \vspace{-0.25cm}
  \caption{\footnotesize{\emph{\textbf{\methodname{} rollout on the takeout-box-packing task.}} In this task, recovery corresponds to driving the gripper backwards and correction corresponds to regrasping the spatula again.}}
  \label{fig:burger_filmstrip}
  \vspace{-0.3cm}
\end{figure}

\newpage

\section{Bimanual Manipulation Tasks Evaluation Protocols}
\label{appendix:task_eval_protocols}
\begin{itemize}
    \item \texttt{shirt-hanging.} Our shirt hanging task setup follows ShirtEasy from ALOHA Unleashed\citep{zhao2024alohaunleashedsimplerecipe} with identical child-size polo shirts, child-size hanger and hanging rack. During policy evaluation, we perform 60 trials per policy. For each of the five shirts, we randomize the initial shirt pose in three orientations (center, left, right) and four hanger placement locations on rack uniformly, resulting in 12 trials per shirt, and 60 trials in total. We also provide an entire uncut evaluation video recording on website \href{https://rac-scaling-robot.github.io/}{https://rac-scaling-robot.github.io/} for reference.
    \item \texttt{airtight-container-lid-sealing.} In this task, we perform 60 trials evaluation for each policy tested. At the beginning of each trial, we randomly assign initial configurations uniformly to 5 different lid placement locations on the drying rack and 12 different container locations on the cutting board, resulting in a total of 60 trials.
    \item \texttt{clamshell-takeout-box-packing.} For this task, we perform 60 trials evaluation for each policy tested. At the beginning of each trial, we randomly assign the locations of the burger uniformly on the right half of the cutting board. For the placement of the takeout box pile and the spatula, we place them roughly in front of the cutting board with a small range of variations each trial.
\end{itemize}

For computing the confidence interval when reporting results and producing the scaling curves, we compute the $95\%$ confidence interval for the task progress scores, where the max scores equal to the maximum number of sub-tasks within each task. For the full task success rates, where each trial receives a binary score for whether the robot completed the entire task successfully, we compute the $95\%$ Wilson score interval, i.e. a formula for binomial proportion confidence interval.

\newpage

\section{Comparison to Prior Works on the Shirt-Hanging Task}
\label{appendix:shirt_comparison}

\textbf{ALOHA Unleashed.} In ALOHA Unleashed \citep{zhao2024alohaunleashedsimplerecipe}, the shirt-hanging task is performed with bimanual ALOHA robot~\citep{zhao2023learning} at two difficulty levels: ShirtEasy and ShirtMessy. ShirtEasy uses 5345 full trajectories and ShirtMessy uses 3313 full trajectories, with a fleet of robots and expert teleoperators. In our work, the shirt-hanging task is designed to be as close to ShirtEasy as possible. They report a full task success rate of $75\%$ on the ShirtEasy task with Diffusion Policy trained on both the ShirtEasy and ShirtMessy data. To standardize the comparison of the size of the data between different works, we approximate the length of the ShirtEasy dataset in \textbf{hours} from ALOHA Unleashed by using an average of $1$ minutes per trajectory. Thus, we estimate a total of $5345 * 60 / 3600 \approx 89$ hours for the ShirtEasy dataset. 

\textbf{Seed GR-3.} In Seed GR-3 \cite{cheang2025gr3technicalreport}, the shirt-hanging task is performed on a custom-designed bimanual mobile manipulation platform. The task differs from ours and \citep{zhao2024alohaunleashedsimplerecipe} in the final step, where the robot ``needs to rotate its mobile base from the table to the drying rack to hang the clothes'', while other sub-tasks remain largely consistent. Importantly, Seed GR-3 reports their performance in \textbf{average task progress}, where a full success corresponds to $1.0$ or $100\%$ and successful completion of each sub-task contributes a fractional score towards the overall task progress. This is different from the \textbf{success rate} metric (Table \ref{tab:prior}), where only full success trials are given score of $1.0$ and other trials do not receive any partial credit. To standardize the evaluation metrics, since ALOHA Unleashed\citep{zhao2024alohaunleashedsimplerecipe} does not report task progress scores, we estimate the full task success rate for GR-3\citep{cheang2025gr3technicalreport} using the Sankey diagram displayed in Figure 10 of their paper, by dividing the vertical heights of the bar representing the last sub-task by the vertical height of the figure location representing the start. This results in a ratio of $7/11 \approx 0.636$.

\end{document}